%% file: main.tex
\documentclass[lettersize,journal]{IEEEtran}
\usepackage{amsmath,amsfonts}
\usepackage{algorithmic}
\usepackage{algorithm}
\usepackage{array}
\usepackage[caption=false,font=normalsize,labelfont=sf,textfont=sf]{subfig}
\usepackage{textcomp}
\usepackage{stfloats}
\usepackage{url}
\usepackage{verbatim}
\usepackage{graphicx}
\usepackage{cite}

\renewcommand{\algorithmicrequire}{ \textbf{Input:}} 
\renewcommand{\algorithmicensure}{ \textbf{Output:}} 
\usepackage{pdfpages}
\usepackage{multirow}
\usepackage[export]{adjustbox} 
\usepackage{bm}
\usepackage{amsmath}
\usepackage{array}
\usepackage{booktabs}
\usepackage{footmisc}
\usepackage{tabularx}

\usepackage{amsthm}
\usepackage{graphicx}
\usepackage{enumitem}
\theoremstyle{definition}

\theoremstyle{Definition}
\newtheorem{mydef}{Definition}
\input{math_commands.tex}

\usepackage{hyperref}
\usepackage{url}
\usepackage{color}
\usepackage{xcolor}
\usepackage{colortbl}
\usepackage{xspace}

\newcommand{\model}{G3AD\xspace}
\begin{document}

\title{Guarding Graph Neural Networks for \\Unsupervised Graph Anomaly Detection}


\author{Yuanchen Bei,
        Sheng Zhou\IEEEauthorrefmark{1},
        Jinke Shi,
        Yao Ma,
        Haishuai Wang,
        and Jiajun Bu
\IEEEcompsocitemizethanks{
    \IEEEcompsocthanksitem Received 25 April 2024; revised 20 February 2025; accepted 4 May 2025. This work was supported in part by the National Natural Science Foundation of China under Grant No. 62476245 and Zhejiang Provincial Natural Science Foundation of China Grant No. LTGG23F030005. \textit{($*$ Corresponding author: Sheng Zhou.)}
    \IEEEcompsocthanksitem Yuanchen Bei, Jinke Shi, and Haishuai Wang are with the College of Computer Science and Technology, Zhejiang University, Hangzhou, China. E-mail: yuanchenbei@zju.edu.cn; shijinke@zju.edu.cn; haishuai.wang@zju.edu.cn.
    \IEEEcompsocthanksitem Sheng Zhou and Jiajun Bu are with the Zhejiang Key Laboratory of Accessible Perception and Intelligent Systems, Hangzhou, China. E-mail: zhousheng\_zju@zju.edu.cn; bjj@zju.edu.cn.
    \IEEEcompsocthanksitem Yao Ma is with the Department of Computer Science, Rensselaer Polytechnic Institute, New York, USA. E-mail: majunyao@gmail.com.
    }
}


\markboth{IEEE TRANSACTIONS ON NEURAL NETWORKS AND LEARNING SYSTEMS}
{Anonymous author \MakeLowercase{\textit{et al.}}: A Sample Article Using IEEEtran.cls for IEEE Journals}


\maketitle

\begin{abstract}
    \input{abstract}
\end{abstract}


\section{Introduction}\label{sec:introduction}
\input{introduction}

\section{Related Works}\label{sec:related}
\input{related.tex}

\section{Preliminaries}\label{sec:pre}
\input{preliminary.tex}

\section{Methodology}\label{sec:method}
\input{method.tex}

\section{Experiments}\label{sec:exper}
\input{experiment.tex}

\section{Conclusion}\label{sec:con}
\input{conclusion}


%

\bibliographystyle{IEEEtran}
\bibliography{IEEEabrv,reference_gad}


\end{document}

%% file: math_commands.tex
\usepackage{amsmath,amsfonts,bm}

\def\eqref#1{equation~\ref{#1}}

\def\1{\bm{1}}

\DeclareMathAlphabet{\mathsfit}{\encodingdefault}{\sfdefault}{m}{sl}
\SetMathAlphabet{\mathsfit}{bold}{\encodingdefault}{\sfdefault}{bx}{n}

\def\gE{{\mathcal{E}}}
\def\gF{{\mathcal{F}}}
\def\gG{{\mathcal{G}}}

\def\gL{{\mathcal{L}}}

\def\gN{{\mathcal{N}}}
\def\gO{{\mathcal{O}}}

\def\gV{{\mathcal{V}}}



%% file: abstract.tex
Unsupervised graph anomaly detection aims at identifying rare patterns that deviate from the majority in a graph without the aid of labels, which is important for a variety of real-world applications.
Recent advances have utilized Graph Neural Networks (GNNs) to learn effective node representations by aggregating information from neighborhoods.
This is motivated by the hypothesis that nodes in the graph tend to exhibit consistent behaviors with their neighborhoods.
However, such consistency can be disrupted by graph anomalies in multiple ways.
Most existing methods directly employ GNNs to learn representations, disregarding the negative impact of graph anomalies on GNNs, resulting in sub-optimal node representations and anomaly detection performance. While a few recent approaches have redesigned GNNs for graph anomaly detection under semi-supervised label guidance, \textit{how to address the adverse effects of graph anomalies on GNNs in unsupervised scenarios and learn effective representations for anomaly detection are still under-explored.}
To bridge this gap, in this paper, we propose a simple yet effective framework for Guarding Graph Neural Networks for Unsupervised Graph Anomaly Detection (\textbf{\model}).
Specifically, \model first introduces two auxiliary networks along with correlation constraints to guard the GNNs against inconsistent information encoding. Furthermore, \model introduces an adaptive caching module to guard the GNNs from directly reconstructing the observed graph data that contains anomalies.
Extensive experiments demonstrate that our \model can outperform twenty state-of-the-art methods on both synthetic and real-world graph anomaly datasets, with flexible generalization ability in different GNN backbones.

%% file: introduction.tex
Graph anomaly detection aims to identify rare patterns or behaviors that significantly deviate from the majority of a graph~\cite{akoglu2015graph,zhao2021synergistic,ma2021comprehensive}. It has attracted increasing attention from both academia and industry due to its practical applications, such as network intrusion detection~\cite{akoglu2015graph}, social spammer detection~\cite{fakhraei2015collective}, and financial fraud detection~\cite{wang2019semi}. The scarcity of anomaly labels in real-world applications, coupled with the challenge of acquiring such labels, has led to a widespread study of unsupervised graph anomaly detection~\cite{duan2023normality,bei2023reinforcement}.
Pioneer works on graph anomaly detection have separately mined the attributed anomalies and topological anomalies by feature engineering and graph structure encoding, respectively~\cite{kim2018encoding,carta2020local}.
Recent breakthroughs in Graph Neural Networks (GNNs) have empowered the unified modeling of attributes and topology, leading to significant advancements in unsupervised graph anomaly detection ~\cite{zhao2020gnn,zhou2021subtractive,zhou2024reconstructed}.

\begin{figure}[tbp]
    \centering
    \includegraphics[width=0.875\linewidth, trim=0cm 0cm 0cm 0cm,clip]{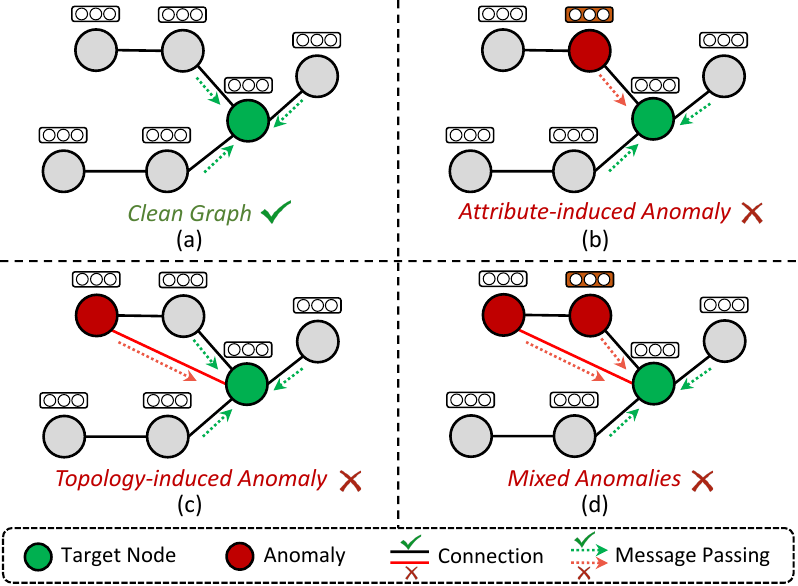}
    \vspace{-0.7em}
    \caption{Toy examples of GNN message passing on clear graphs and the graphs under three types of anomaly impacts.}
    \vspace{-1.4em}
    \label{fig:inconsist_example}
\end{figure}

The success of representative GNNs, as illustrated in Figure \ref{fig:inconsist_example}-(a), has heavily relied on aggregating information from neighborhoods sharing similar patterns, e.g. label homophily and feature consistency~\cite{mcpherson2001birds,ma2022is}. 
Although successful, in graphs with anomalies, we argue that this foundation can be easily undermined by various graph anomalies in multiple ways~\cite{yang2022graph}.
Figure \ref{fig:inconsist_example}-(b-d) illustrates toy examples of the three main types of graph anomaly impacts on GNNs.
(i) As illustrated in Figure \ref{fig:inconsist_example}-(b), the \textbf{attributed-induced anomalies} where node features are corrupted, such as the account takeover in social networks, will directly result in aggregating incorrect information from neighborhoods.
(ii) As illustrated in Figure \ref{fig:inconsist_example}-(c), the \textbf{topological-induced anomalies} that connect to incorrect neighborhoods, such as the fraudsters in E-commerce systems, will result in aggregating information from inconsistent neighborhoods.
(iii) More seriously, as illustrated in Figure \ref{fig:inconsist_example}-(d), the \textbf{mixed anomalies} that simultaneous occurrence of multiple anomaly types is a more common situation and will have a significantly more detrimental impact on GNNs.
Existing unsupervised graph anomaly detection methods have primarily focused on designing effective unsupervised anomaly scoring functions within the representation space learned directly from GNNs while \textit{overlooking the negative impact of anomalies on the inherent GNNs themselves}~\cite{ma2021comprehensive,zheng2021generative,zhang2022subcr}. 
This is crucial for accurate anomaly detection, which is indispensable to discriminative node representations.
Under this limitation, the representation learning ability of GNNs is hampered from adequately capturing normal patterns in the graph, thereby leading to suboptimal representation learning and anomaly detection performance. Hence, it is crucial to alleviate the negative impact of anomalies and fully release the power of GNNs for anomaly detection.

Although important, addressing the negative impacts of graph anomalies on GNNs in unsupervised scenarios is non-trivial.
An intuitive solution might involve the development of innovative GNNs that adapt to certain graph anomalies. This has attracted increasing attention, and the few very recent works have achieved tremendous success in semi-supervised scenarios~\cite{chai2022can,tang2022rethinking}. 
However, the lack of labels in unsupervised anomaly detection poses a significant challenge in guiding the redesign of GNNs, which is more commonly encountered in practice.
More importantly, while anomalies do exist on the graph and exert certain effects on GNNs, they account for only a small fraction of the total. The majority of patterns remain normal and can be effectively captured by GNNs. Therefore, a complete redesign or deprecation of GNNs could risk compromising the identification of the majority of normal patterns due to the influence of a minor proportion of anomalies, especially in unsupervised contexts.
Therefore, \textit{how to reduce the impact of anomalies on GNNs while enabling them to capture the majority of normal patterns effectively and aid in unsupervised anomaly detection?}

\begin{figure}[tbp]
    \centering
    \includegraphics[width=0.845\linewidth, trim=0cm 0cm 0cm 0cm,clip]{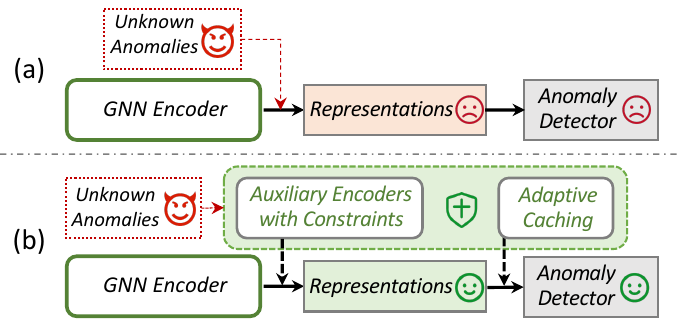}
    \vspace{-0.5em}
    \caption{Concept maps between (a) existing GNN-based unsupervised graph anomaly detection paradigm and (b) our proposed GNN-guarded unsupervised graph anomaly detection paradigm.}
    \vspace{-1.2em}
    \label{fig:example}
\end{figure}

To answer the above under-explored research question, in this paper, we propose a simple yet effective framework for \textbf{\underline{G}}uarding \textbf{\underline{G}}raph Neural Network for Unsupervised \textbf{\underline{G}}raph \textbf{\underline{A}}nomaly \textbf{\underline{D}}etection (\textbf{\model}). 
Instead of directly encoding the observed graph with anomalies, \model first introduces two auxiliary encoders tailored with correlation constraints to guard the GNNs against encoding inconsistent information.
Subsequently, to optimize GNNs in an unsupervised manner and detect multiple types of unknown anomalies, \model proposes a comprehensive learning objective that includes both local attribute/topology reconstruction and global consistency alignment, which also serves as anomaly scoring. 
\textcolor{black}{During this process, to avoid directly reconstructing the observed graph with anomalies \model introduces an adaptive caching module to guard the GNNs from misleading learning objectives during model training.} 
Figure \ref{fig:example} illustrates the comparison between the existing GNN-based unsupervised graph anomaly detection paradigm and the one under \model framework.
\textcolor{black}{Extensive experiments on both four synthetic and two real-world graph anomaly datasets demonstrate that \model outperforms twenty state-of-the-art unsupervised graph anomaly detection models. Further in-depth analysis from diverse perspectives also demonstrates the strengths of \model.}
The main contributions of this paper are summarized as follows:
\begin{itemize}[leftmargin=*]
    \item We emphasize the negative impact of unknown anomalies on GNNs, which is crucial for graph anomaly detection under unsupervised settings while overlooked by most existing unsupervised works.
    \item We propose \model, a simple yet effective framework to guard GNNs from encoding inconsistent information and directly reconstructing the abnormal graph in unsupervised graph anomaly detection.
    \item We conduct extensive experiments on twenty widely-used datasets, including both synthetic and real-world scenarios. Experimental results show that the proposed \model outperforms twenty state-of-the-art baselines.
\end{itemize}

In the following sections, we will first review the previous work related to our method in Section \ref{sec:related}. Second, we will give some key preliminaries of our work in Section \ref{sec:pre}.
Then, the detailed description of the proposed \model will be introduced in Section \ref{sec:method}.
To further verify the effectiveness of \model, we conduct various experiments in Section \ref{sec:exper}.
Finally, the conclusion of this paper is posed in Section \ref{sec:con}.

%% file: related.tex
\subsection{Graph Neural Networks}\label{sec:related_gnn}
Graph Neural Networks (GNNs) are a series of deep learning models specifically designed for processing graph-structured data~\cite{kipf2016semi,hamilton2017inductive}. The core design of GNNs is to learn node representations by aggregating and propagating information from neighborhoods to the central nodes. This information propagation (message passing) allows GNNs to capture complex topology dependencies and contextual information among nodes~\cite{zhang2020deep,wu2020comprehensive}.

Typically, in GNNs, each node has an initial feature representation, and through multiple layers of message aggregation and propagation operations, the node representations are gradually updated and refined~\cite{kipf2016semi,hamilton2017inductive,veličković2018graph}.
Representatively, GCN~\cite{kipf2016semi} adopts the graph convolution operator and stacks it into multiple layers for neighbor message passing. GAT~\cite{veličković2018graph} further introduces the attention mechanism to consider the different importance of neighbor nodes and dynamically assign different weights when aggregating neighbor features. GraphSAGE~\cite{hamilton2017inductive} extends and improves the graph convolution to handle large-scale graph data, with the key idea of generating node representations through sampling and aggregating neighbor nodes.

Due to the fact that real-world data can be widely modeled as graphs, GNNs have emerged as a significant branch within the field of neural network research~\cite{dong2021survey,wu2020comprehensive}. They possess the ability to perform learning on graph-structured data, automatically extracting features and making predictions, providing an effective solution for tasks on graph data.
Based on these advantages, GNNs have achieved significant success in various application domains, such as social network analysis~\cite{kumar2022influence,jain2023opinion}, recommendation systems~\cite{he2020lightgcn,chen2024macro}, and bioinformatics~\cite{zhang2021graph,yi2022graph}.

\subsection{Unsupervised Graph Anomaly Detection}
Graph anomaly detection technologies have been widely applied in real-world systems to ensure their robustness and security, such as financial transaction networks~\cite{tang2022rethinking}, social network applications~\cite{yu2016survey}, and e-commerce systems~\cite{deldjoo2021survey}.

Existing graph anomaly detection methods in unsupervised manners can be largely divided into four main categories.
(i) \textit{Classical shallow models}: SCAN~\cite{xu2007scan} can be utilized for anomaly detection based on structural similarity. MLPAE~\cite{sakurada2014anomaly} detects anomalies with nonlinear transformations.
(ii) \textit{Enhanced graph neural networks}: GAAN~\cite{chen2020generative} utilizes a generative adversarial framework to train the graph encoder with real graph and anomalous fake graph samples. ALARM~\cite{peng2020deep} proposes a multi-view representation learning framework with multiple graph encoders and a well-designed aggregator. AAGNN~\cite{zhou2021subtractive} then enhances the graph neural network with an abnormality-aware aggregator. 
(iii) \textit{Deep graph autoencoder models}: Dominant~\cite{ding2019deep} first introduces a deep graph autoencoder model with a shared encoder to measure anomalies by reconstruction error.
AnomalyDAE~\cite{fan2020anomalydae} introduces asymmetrical cross-modality interactions between autoencoders. 
ComGA~\cite{luo2022comga} further designs a community-aware tailored graph autoencoder to make the representation between normal and anomalous nodes more distinguishable. 
\textcolor{black}{Recently, CoCo~\cite{wang2024context} enhances the graph autoencoder-based anomaly detector with node-level context correlation modeling.} 
(iv) \textit{Graph contrastive learning methods}: CoLA~\cite{liu2021anomaly} exploits the local information and introduces a self-supervised graph contrastive learning method to detect anomalies.
Further, ANEMONE~\cite{jin2021anemone} utilizes the graph contrastive learning method at multiple scales.
SL-GAD~\cite{zheng2021generative} further performs anomaly detection from both generative and multi-view contrastive perspectives.
Sub-CR~\cite{zhang2022subcr} then proposes a self-supervised method based on multi-view contrastive learning with graph diffusion and attribute reconstruction.
\textcolor{black}{Futrher, ARISE~\cite{duan2023arise} enhances graph contrastive learning with substructure discovery. NLGAD~\cite{duan2023normality} adopts multi-scale graph contrastive learning with a hybrid normality measurement.} 

However, the above GNN-based models directly apply the message passing without anomaly guarding, 
\textcolor{black}{neglecting the impact of neighborhood inconsistency on GNN mechanisms, thus the performance of GNNs is harmed and limited.}

\subsection{Neighborhood Consistency in Graphs}
The nature that neighboring nodes in real-world graphs tend to share consistent behaviors, such as labels or attributes~\cite{wang2022powerful}, has been the foundation of many GNNs.
As introduced in Section \ref{sec:related_gnn}, pioneer representative works of GNNs\cite{kipf2016semi,hamilton2017inductive,veličković2018graph} have utilized the neural message passing by aggregating information from neighbors, which is a straightforward way of applying the feature consistency.
Later works have proved that such message passing is equivalent to classic label propagation, where the labels are propagated along with the connection between neighboring nodes~\cite{wang2021combining}.
This can be viewed as utilizing the neighborhood consistency in labels, which is also widely called the homophily assumption in practice~\cite{ma2022is}.

However, the homophily assumption may not always hold in real-world graphs~\cite{zhu2021graph,wu2023homophily,chen2023exploiting}.
To tackle this challenge, recent works have been made on the \textit{heterophily GNN} for representation learning under graphs with low neighborhood consistency. 
Representatively,
MixHop~\cite{abu2019mixhop} mixes powers of the adjacency matrix for graph convolution to ease the limitation.
H2GCN~\cite{zhu2020beyond} designs a model with ego \& neighbor separation, higher-order neighbors, and intermediate representations combination.
LINKX~\cite{lim2021large} separately embeds the adjacency and node features with simple MLP transformations rather than the aggregation based on neighborhood consistency.
GloGNN~\cite{li2022finding} performs aggregation from the whole set of nodes with both low-pass and high-pass filters.

Nevertheless, in unsupervised graph anomaly detection, these heterophily GNNs mentioned above are not directly appropriate, due to the unknown neighborhood inconsistency in attribute and topology under the unsupervised setting and the overlook of the anomaly-specific design.

%% file: preliminary.tex
In this section, we present some key notations and definitions related to our target unsupervised graph anomaly detection task. 
Note that we focus on node-level graph anomaly detection in this paper.
For the convenience of readers, we list the main symbols used in this paper in Table \ref{tab:symbol}.

\begin{table}[tbp]
  \centering
  \caption{Key symbols and definitions in this paper.}
  \resizebox{\linewidth}{!}{
    \begin{tabular}{c|c}
    \toprule
    \multicolumn{1}{c|}{Notations} & Descriptions \\
    \midrule
    \midrule
     $\mathcal{G}=(\bm{A}, \bm{X})$     & An attributed graph with anomalies. \\
     \hline
     $\bm{A}$ &  The graph adjacency matrix. \\
     \hline
     $\bm{X}$ &  The node attribute matrix. \\
     \hline
     $\mathcal{V}$ & The set of nodes in $\mathcal{G}$. \\
     \hline
     $\mathcal{E}$     & The set of edges in $\mathcal{G}$. \\
     \hline
     $\gN_{i}$ & The neighborhood set of node $v_{i}$. \\
     \hline
     $n=|\gV|$ & The number of nodes in $\mathcal{G}$. \\
     \hline
     $d$ & The dimension of node attributes in $\mathcal{G}$. \\
     \hline
     $k$ & The number of graph anomalies in $\mathcal{G}$.\\
     \hline
     $\bm{Y}$ & The anomaly label set of nodes. \\
     \hline
     $\bm{S}$ & The anomaly score vector indicating node abnormalities. \\
     \midrule
    \bottomrule
    \end{tabular}%
    }
  \label{tab:symbol}%
\end{table}%

\noindent \textbf{Notations.} Let $\gG = (\bm{A}, \bm{X})$ be a graph with the node set $\gV = \{v_{1}, v_{2}, ..., v_{n}\}$ and the edge set $\gE$, where $|\gV| = n$. $\bm{A} \in \mathbb{R}^{n \times n}$ denotes the graph adjacency matrix, for each element ${A}_{i,j}$ in $\bm{A}$, ${A}_{i,j} = 1$ indicates that there is an edge between node $v_{i}$ and node $v_{j}$, and otherwise $A_{i,j} = 0$. $\bm{X} \in \mathbb{R}^{n \times d}$ denotes the node attribute matrix, where the $i$-th row vector $\bm{x}_{i} = \bm{X}[i,:] \in \mathbb{R}^{d}$ indicates the attribute vector of $v_{i}$ with $d$ dimensional representation.
$\gN_{i}$ is the neighborhood set of a central node $v_i$ in the graph $\gG$.

\begin{mydef}
    \textbf{Graph Neural Networks}: In form, for an $L$-layer GNN, the calculation process of each layer can be expressed as the aggregation and updating operators. In the aggregation phase, each central node aggregates the message from its neighbor nodes:
    \begin{equation}
        \bm{m}_{i}^{(l+1)} = \text{Aggregator}(\{\bm{h}_{j}^{(l)}|j\in \gN_{i}\}),
    \end{equation}
    where the function $\text{Aggregator}(\cdot)$ is the message aggregation operator to aggregate information from nodes' neighborhoods, $\bm{h}_{j}^{(l)}$ is the representation of node $v_{j}$ in the $l$-th GNN layer, and $\bm{m}_{i}^{(l+1)}$ is the aggregated message for node $v_{i}$ in the $(l+1)$-th GNN layer. After obtaining the aggregated message, in the updating phase, each central node adopts the aggregated information to update and transform its own representations:
    \begin{equation}
        \bm{h}_{i}^{(l+1)} = \text{Updater}(\bm{h}_{i}^{(l)}, \bm{m}_{i}^{(l+1)}),
    \end{equation}
    where the function $\text{Updater}(\cdot)$ is the node representation updating operator and $\bm{h}_{i}^{(l+1)}$ is the updated representation of node $v_{i}$.
\end{mydef}

\begin{mydef}
    \textcolor{black}{\textbf{Graph Anomalies}: Given an abnormal attributed graph $\gG = (\bm{A}, \bm{X})$ containing $n$ node instances, and $k$ of them are graph anomalies ($k \ll n$), whose attributes, connections or behaviors are different from most other normal nodes. On a graph with anomalies, each node $v_i$ is associated with an anomaly label $y_{i}\in \bm{Y}$, where $\bm{Y}$ denotes the anomaly label set and each element $y_{i} \in \{0,1\}$ denotes whether node $v_{i}$ is an anomaly.}
\end{mydef}

\begin{mydef}
    \textcolor{black}{\textbf{Unsupervised Graph Anomaly Detection}: Given an abnormal graph as Definition 2, the target of unsupervised graph anomaly detection is to learn a model $\gF(\cdot): \mathbb{R}^{n \times n} \times \mathbb{R}^{n \times d} \rightarrow \mathbb{R}^{n}$ in an unsupervised manner that outputs anomaly score vector $\bm{S}\in \mathbb{R}^{n}$ to measure the degree of abnormality of nodes. Specifically, the $i$-th element $s_i$ in the score vector $\bm{S}$ indicates the abnormality of node $v_{i}$, where a larger score means a higher abnormality. Note that the anomaly label set $\bm{Y}$ is invisible in the unsupervised setting.}
\end{mydef}

%% file: method.tex
\begin{figure*}[t]
    \centering
    \includegraphics[width=\linewidth, trim=0cm 0cm 0cm 0cm,clip]{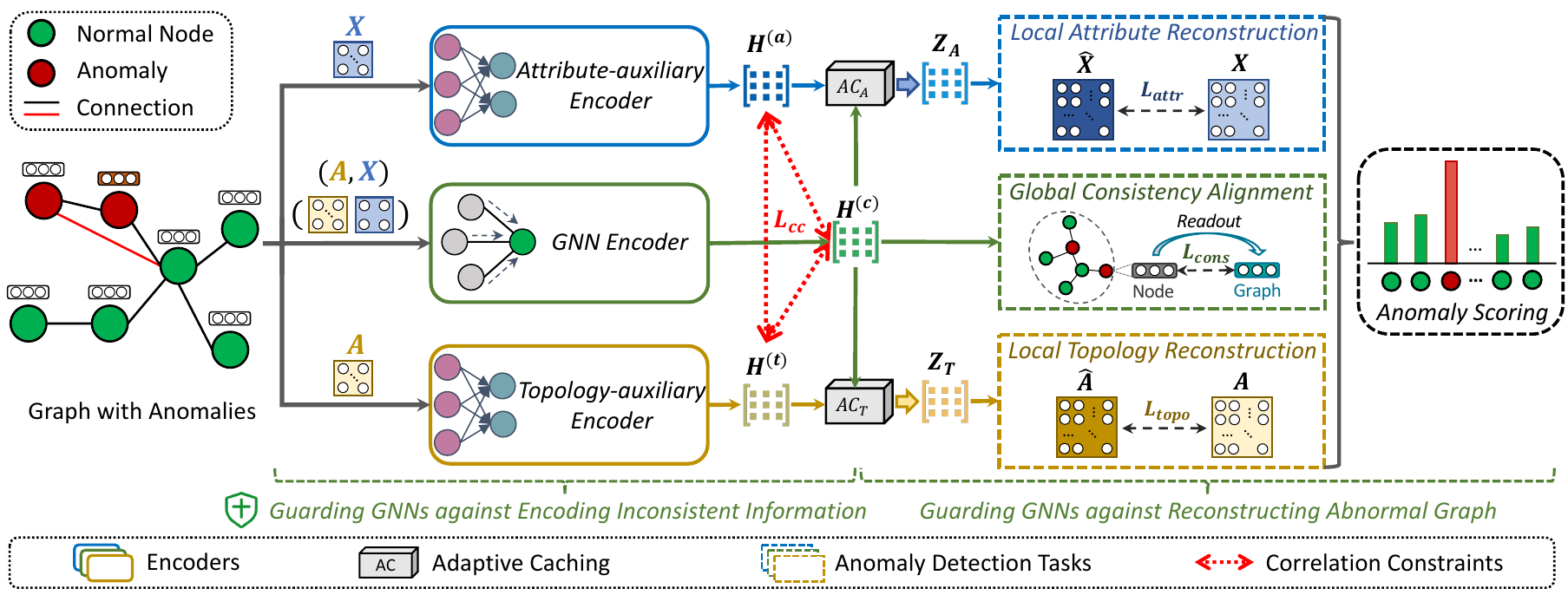}
    \caption{The overall architecture of \model guarding framework contains two major novel parts: (i) Guarding GNNs against encoding inconsistent information with two auxiliary encoders with correlation constraints; (ii) Guarding GNNs against reconstructing abnormal graphs with adaptive information caching. Under the two guards, we comprehensively consider both local reconstruction and global alignment to detect different types of anomalies.}
    \label{fig:framework}
\end{figure*}

In this section, we present the details of the Guarding Graph Neural Network for Unsupervised Graph Anomaly Detection (\model). The overall framework of \model is illustrated in Figure \ref{fig:framework}. 
Generally, \model follows a representative GNN-based anomaly detection scheme where the GNNs learn the representations, and the representations are optimized with unsupervised objectives and measured for anomaly detection.
Besides, \model introduces two novel guarding strategies to the paradigm, namely guarding against encoding inconsistent information and guarding against reconstructing abnormal graphs, so that the GNNs can produce effective representations and boost the anomaly detection ability. With carefully designed guarding strategies and equipped anomaly detection tasks, \model can outperform current GNN-based models.
In the rest of this section, we will introduce the details of these guarding strategies and how \model serves unsupervised graph anomaly detection.

\subsection{Guarding GNNs against Encoding Inconsistent Information}

\textbf{Motivation}. As discussed in previous works, GNNs are designed for capturing consistent information from neighborhoods~\cite{yang2022graph}. However, in graphs with anomalies, the anomaly-induced inconsistent patterns in unsupervised graph anomaly detection have been largely overwhelmed and will negatively disrupt the GNN performance.
Therefore, \model first introduces two auxiliary encoders with correlation constraints to design an encoding guarding strategy for GNNs.
Specifically, the auxiliary encoders are expected to encode the inconsistent information from the attribute and topology perspectives, and the GNNs are guarded against encoding the inconsistent information under the correlation constraints among the representations learned by the three encoders.

\subsubsection{GNN Encoder} 
The GNN encoder is utilized for encoding the part of the information that satisfies the consistent homophily assumption, which is \textbf{shared} by both attribute and topology. 
To focus on the model architecture, we adopt a two-layer Graph Attention Network~(GAT)~\cite{veličković2018graph} as the GNN encoder $f_{gnn}(\cdot)$ for simplicity. It is worth noting that any other GNNs can be directly applied here.
Given the input attributes $\bm{x}_i$ and $\bm{x}_j$ of node $v_i$ and its neighbor $v_j$, a GAT layer learns the input attributes with the attention mechanism. The attention coefficient is computed as:
\begin{equation}
    e_{i,j}=att(\bm{x}_i, \bm{x}_j)=LeakyReLU(\bm{a}^{T}\cdot[\bm{W}_{enc}\bm{x}_i||\bm{W}_{enc}\bm{x}_j]),
\end{equation}
where $att(\cdot)$ is a single-layer feedforward neural network, parameterized by weight vector $\bm{a}\in\mathbb{R}^{2d'}$ and $\bm{W}_{enc}\in\mathbb{R}^{d' \times d}$ with the transformed dimension $d'$, $||$ denotes the concatenate operation. The attention coefficient $\alpha_{i,j}$ is normalized as:
\begin{equation}
    \alpha_{i,j}=Softmax(e_{i,j})=\frac{exp(e_{i,j})}{\sum_{k \in \gN_{i}}exp(e_{i,k})},
\end{equation}
where $\gN_{i}$ denotes the neighbor set of node $v_{i}$.
Then, the final output feature $\bm{h}_{i} \in \mathbb{R}^{d'}$ of node $v_{i}$ can be obtained as:
\begin{equation}
    \bm{h}_{i}=\sum_{j \in \gN_{i}}\alpha_{i,j}\bm{W}_{enc}\bm{x}_j.
\end{equation}
For convenience, we use $\bm{H}^{(c)} \in \mathbb{R}^{n \times d'}$ denote the output representation of consistency encoder $f_{gnn}(\cdot)$.
\textcolor{black}{Note that \model can be generalized in different GNN backbones, and any other graph message passing functions~\cite{hamilton2017inductive,xu2018gin} can also be adopted here. We will discuss the generalization ability of \model in Section~\ref{sec:general_exp}.}

\subsubsection{Auxiliary Encoders with Correlation Constraints}
In our \model, the attribute-auxiliary encoder and the topology-auxiliary encoder are further utilized for encoding the part of information dependent on each information source~(attribute and topology) as well as the potential specific type of anomalies under the attribute/topology inconsistent situation.

Existing methods directly utilize GNNs to model the attribute and topology simultaneously. 
\textcolor{black}{However, in the part where the consistency is destroyed by anomalies, message passing in GNNs will introduce negative noise information, even helping anomalies to \textit{camouflage} when passing normal messages to them~\cite{chai2022can}. Therefore, here we use two \textbf{independent} MLPs without graph message passing, rather than GNN encoders based on neighborhood consistency, to better model the inconsistent attribute/topology patterns and guard the GNN encoder learning.}

Specifically, the two independent MLPs $f_{a}(\cdot)$ and $f_{t}(\cdot)$ are applied to encode the original attribute matrix $\bm{X}$ and the adjacency matrix $\bm{A}$, respectively:
\begin{equation}
    \bm{H}^{(a)} = f_{a}(\bm{X}),\quad \bm{H}^{(t)} = f_{t}(\bm{A}),
\end{equation}
where $f_{a}(\cdot)$, $f_{t}(\cdot)$ are two MLPs with learnable parameters $\bm{W}_{a}$ and $\bm{W}_{t}$ respectively, and $\bm{H}^{(a)}, \bm{H}^{(t)} \in \mathbb{R}^{n \times d'}$ are the corresponding output matrices. 

To ensure better guarding of the GNN encoder against those inconsistent patterns, it is desired that the representations encoded from the GNN and the auxiliary encoders are more independent of each other. 
Therefore, we impose the correlation constraint on the embedding space of the encoded representation to ensure all aspects of information $\bm{H}^{(a)}$, $\bm{H}^{(t)}$, and $\bm{H}^{(c)}$ are well-disentangled encoding.

Specifically, the correlation constraint between each pair of encoded representation aspects is designed as the absolute correlation coefficient measurement. Minimizing the constraint ensures the two paired vectors become more independent~\cite{asuero2006correlation}. Take the constraint between $\bm{H}^{(a)}$ and $\bm{H}^{(t)}$ as an example, the formal expression is as follows:
\begin{equation}
    aCor(\bm{H}^{(a)}, \bm{H}^{(t)}) = abs\left(\frac{Cov(\bm{H}^{(a)}, \bm{H}^{(t)})}{\sqrt{Var(\bm{H}^{(a)}) \cdot Var(\bm{H}^{(t)})}}\right),
\end{equation}
\textcolor{black}{where $Cov(\cdot)$ is the covariance between two matrices, and $Var(\cdot)$ is a matrix's own variance, and $abs(\cdot)$ is the absolute value function. $aCor(\bm{H}^{(a)}, \bm{H}^{(c)})$ and $aCor(\bm{H}^{(t)}, \bm{H}^{(c)})$ can be calculated in the same way. $Cov(\cdot)$ is insensitive to the order of the two inputs.} 
The overall regularization $\gL_{cc}$ of all correlation constraints can be presented as:
\begin{equation}
\begin{aligned}
    \gL_{cc} = & aCor(\bm{H}^{(a)}, \bm{H}^{(c)}) + aCor(\bm{H}^{(t)}, \bm{H}^{(c)}) \\&+ aCor(\bm{H}^{(a)}, \bm{H}^{(t)}),
\end{aligned}
\label{eq:cc}
\end{equation}
where the calculated $\gL_{cc}$ is utilized as a correlation regularization loss during the model training stage.

\subsection{Guarding GNNs against Reconstructing Abnormal Graph}
\textbf{Motivation}. 
Graph anomalies have various types, including attribute anomalies, topology anomalies, and mixed anomalies. In unsupervised graph anomaly detection, we are unable to know either anomaly labels or anomaly types. Thus, \model aims to comprehensively detect different anomaly types in the following ways:
\begin{itemize}[leftmargin=*]
    \item \textbf{Local attribute reconstruction}: the attribute reconstruction errors distinguishing sparse attribute anomalies from the predominant normal nodes. 
    \item \textbf{Local topology reconstruction}: the topology reconstruction errors distinguishing sparse topology anomalies from the primary normal nodes. 
    \item \textbf{Global consistency alignment}: the alignment distance between node representations simultaneously considering attribute and topology information encoded by the GNN and the global graph consistency vector can further be equipped for anomaly distinguishing.
\end{itemize}
Among these objectives, the graph reconstruction scheme has been proven to be essential for both representation optimization and anomaly detection under the unsupervised setting~\cite{ding2019deep,fan2020anomalydae,roy2024gad}.
\textcolor{black}{However, reconstruction targets are needed to make GNN-encoded consistent representations to fit the observed graph data with anomalies. The unknown anomalies in the abnormal graph may provide a misleading objective to GNNs for both tasks.}
Thus, \model further introduces adaptive caching to guard the GNNs against reconstructing abnormal graphs by cooperating with the auxiliary encoders that partake the inconsistent anomaly patterns.

\subsubsection{Adaptive Caching}
\textit{\textbf{From the consistent alignment perspective}}, since the consistent information encoded by the GNN encoder has already been guarded, it should be directly used for measuring the global consistency under a mixture of attribute and topology information. 
\textit{\textbf{Yet from the attribute/topology reconstruction perspective}}, due to the graph to be reconstructed containing anomalies, it is necessary to further guard the GNN-encoded representations against directly reconstructing and preserving the abnormal graph. 

Therefore, we design an Adaptive Caching (AC) module to concurrently leverage the GNN-encoded representations for normal part reconstruction and auxiliary-encoded representations for inconsistent part reconstruction, avoiding the force of GNNs to fit inconsistent anomaly patterns.
It automatically selects appropriate information for the reconstruction of normal and abnormal parts from GNN representation and auxiliary representation under unsupervised conditions, through learnable parameters.
The general AC module can be formulated as follows:
\begin{gather}
    \bm{imp} = Tanh(\tau(\bm{H}^{(1)}||\bm{H}^{(2)})),\\
    \text{AC}(\bm{H}^{(1)},\bm{H}^{(2)}) = \bm{imp}_{[:n]} \cdot \bm{H}^{(1)} + \bm{imp}_{[n:]} \cdot \bm{H}^{(2)},
\end{gather}
where $\tau(\cdot)$ is an MLP for information fusion, $\bm{imp} \in \mathbb{R}^{2n}$ is the weight vector of each dimension of the input features, $\bm{H}^{(1)}$ and $\bm{H}^{(2)}$ denote two information sources, such as $\bm{H}^{(a)}$, $\bm{H}^{(t)}$, and $\bm{H}^{(c)}$.

The AC module is utilized to generate two types of representations for attribute and topology reconstruction:
\begin{equation}
    \bm{Z}_{A} = \text{AC}(\bm{H}^{(a)},\bm{H}^{(c)}),\quad \bm{Z}_{T} = \text{AC}(\bm{H}^{(t)},\bm{H}^{(c)}),
\end{equation}
where $\bm{Z}_{A}$ and $\bm{Z}_{T}$ are the cached attribute representation and cached topology representation.
Compared with vanilla GNN-based representation, the caching of GNN-encoded and attribute/topology-specific information automatically selects helpful information and better alleviates anomalies’ impact on the GNN-encoded representations.

\subsubsection{Anomaly Detection Tasks}
Given the GNN encoded representation, cached attribute representation, and cached topology representation, we aim to make full use of them for anomaly scoring in an unsupervised setting.
Therefore, as the above definition of the three types of anomaly detection objectives, the detection tasks can be unfolded as follows.

\textbf{Local attribute reconstruction:}
The reconstruction-based operator has been widely observed to be effective for unsupervised anomaly detection~\cite{ding2019deep,bei2023reinforcement,roy2024gad}.
Therefore, given the cached attribute representation $\bm{Z}_{A}$, we utilize the node's attribute reconstruction ability for attribute anomaly mining, benefiting from the reconstruction target tending to fit the patterns of the predominant nodes while the patterns of anomalies are rare~\cite{ding2019deep}.
We use a two-layer Graph Convolution Network~(GCN)~\cite{kipf2016semi} as the attribute reconstruction function $g_a(\cdot)$ as follows:
\begin{equation}
    \bm{R}^{(l+1)} = LeakyReLU(\tilde{\bm{A}}\bm{R}^{(l)}\bm{W}_{re}^{(l)}),
\end{equation}
where $\tilde{\bm{A}}=\hat{\bm{D}}^{-\frac{1}{2}}\bm{A}^{*}\hat{\bm{D}}^{-\frac{1}{2}} \in \mathbb{R}^{n \times n}$ is the normalized adjacency matrix, $\hat{\bm{D}} \in \mathbb{R}^{n \times n}$ is the degree matrix of $\bm{A}^{*}=\bm{A}+\bm{I}$ where $\bm{I}$ is the identity matrix. $\bm{R}^{(l)}$, $\bm{W}_{re}^{(l)}$ is the input features, trainable parameters in $l$-th layer, respectively.

For the cached attribute representation $\bm{Z}_{A}$, $g_{a}(\cdot)$ reconstructs the attribute matrix $\hat{\bm{X}} \in \mathbb{R}^{n \times d}$ with corresponding distance-based reconstruction loss $\gL_{attr}$, which can be computed as:
\begin{gather}
    \hat{\bm{X}}=g_{a}(\bm{Z}_{A}, \bm{A}),\\
    \gL_{attr} = ||\bm{X}-\hat{\bm{X}}||_{2}^{2},
\label{eq:attr}
\end{gather}
where $||\cdot||_{2}^{2}$ is the Euclidean distance.

\textbf{Local topology reconstruction:}
Similar to the perspective of local attribute reconstruction, for the given cached topology representation $Z_{T}$, the topology reconstruction capability is also a factor for anomaly detection due to topology anomalies' rare and inconsistent local topological patterns.
The reconstruction function $g_{t}(\cdot)$ reconstructs the adjacency matrix from $\bm{Z}_{T}$ with the reconstruction loss $\gL_{topo}$ as follows:
\begin{gather}
    \hat{\bm{A}}=g_{t}(\bm{Z}_{T})=\bm{Z}_{T}\cdot\bm{Z}_{T}^{'},\\
    \gL_{topo} = ||\bm{A}-\hat{\bm{A}}||_{2}^{2},
\label{eq:topo}
\end{gather}
where $g_{t}(\cdot)$ is designed as an inner product between $\bm{Z}_{T}$ and its self-transposition $\bm{Z}_{T}^{'}$ to make the reconstruction efficient.

\textbf{Global consistency alignment:}
\textcolor{black}{Given the GNN encoded representation $\bm{H}^{(c)}$ with both the attribute and topology information by graph message passing to detect anomalies, a naive way for evaluation is to measure the distance between node embeddings and their corresponding subgraph. However, the anomalies may occur as neighbors of normal nodes under the neighborhood inconsistency. Aligning the node with a noisy subgraph may be suboptimal. Thus, due to abnormal nodes making up only a minority of the entire graph, we turn to measure each node embedding with the graph summary vector for global consistency measurement to reduce the negative impact of neighborhood on the summary vectors.}

Specifically, to conduct the alignment, we first read out the consistent GNN-encoded representations of nodes into a graph summary representation:
\begin{equation}
    \bm{E}_{g}=\text{Readout}(\bm{H}^{(c)}),
\label{eq:readout_func}
\end{equation}
where $\text{Readout}(\cdot)$ can be a kind of pooling operation (such as min, max, mean, and weighted pooling~\cite{liu2021anomaly}), here we use the mean pooling as default for simplicity, $\bm{E}_{g}$ is the readout representation of the graph.

Then we conduct the global consistency alignment task between each node's GNN-encoded representation $\bm{z}_{c,i} \in \bm{H}^{(c)}$ and the graph summary vector $\bm{E}_{g}$ as follows:
\begin{equation}
    \gL_{cons}=log(\sqrt{\sum_{i=1}^{n}||\bm{z}_{c,i}-\bm{E}_{g}||_{2}^{2}}+e),
\label{eq:homo}
\end{equation}
where $\bm{z}_{c,i}$ is the consistent representation of node $v_{i}$, constant $e$ is used to limit the lower bound of the loss for better balance $\gL_{cons}$'s numerical relationship with other loss terms.

\subsection{Anomaly Scoring}
With the three different anomaly detection tasks, we utilize the above three anomaly detection tasks for the final anomaly scoring.
The normal nodes in the graph are expected to show low discrepancies, while anomalies exhibit high discrepant values due to their inconsistency, irregularity, and diversity.
Therefore, here we compute the anomaly score $s_{i}$ of each node $v_i$ in multi-perspectives according to:
\begin{equation}
\begin{aligned}
    s_{i}=&\underbrace{\lambda_{1}\cdot||\bm{x}_{i}-\hat{\bm{x}}_{i}||_{2}^{2}}_{\text{Local Attribute Reconstruction}} + \underbrace{(1-\lambda_{1})\cdot ||\bm{a}_{i}-\hat{\bm{a}}_{i}||_{2}^{2}}_{\text{Local Topology Reconstruction}}\\&+\underbrace{\lambda_{2}\cdot log(\sqrt{||\bm{z}_{c,i}-\bm{E}_{g}||_{2}^{2}}+e)}_{\text{Global Consistency Alignment}},
\end{aligned}
\label{eq:scoring}
\end{equation}
where $\lambda_{1}$ is a hyperparameter to balance the attribute and topology reconstruction, and $\lambda_{2}$ is also a hyperparameter that measures the effect of the global consistency alignment.
Nodes with larger scores are more likely to be considered as anomalies, thus we can compute the ranking of anomalies according to the nodes' scores calculated as Eq.(\ref{eq:scoring}).

\subsection{Joint Training Objective Function}
\textcolor{black}{Following previous works to optimize the model in the absence of labels~\cite{ding2019deep,liu2021anomaly}, to jointly train different aspects of anomaly scoring loss with the correlation constraint for GNN guarding, the training objective function of \model:}
\begin{equation}
    \gL = \lambda_{1}\cdot{\gL}_{attr} + (1-\lambda_{1})\cdot{\gL}_{topo} + \lambda_{2}\cdot{\gL}_{cons} + \gL_{cc},
\label{eq:hyper}
\end{equation}
where the hyperparameters $\lambda_{1}$ and $\lambda_{2}$ here are the same as mentioned in Eq.(\ref{eq:scoring}).
In this way, \model is trained using the gradient descent algorithm on the overall joint training objective function.

Conclusively, the pseudocode of the overall procedure workflow of \model is described as the given Algorithm \ref{alg:g3ad}. In each training epoch, \model first encodes three aspects of representations by the guarded GNN encoder along with the auxiliary encoders. Further, we adopt the correlation constraint to minimize correlations between the three representations. Before conducting the anomaly detection tasks, the adaptive caching module is equipped for GNN representation guarding. Then, anomaly detection tasks are conducted for anomaly scoring and joint objective loss obtaining. A backpropagation is then executed with a gradient descent algorithm to optimize the parameters of \model. Finally, the anomaly scores are returned for each node to evaluate their abnormality.

\begin{algorithm}[tb]
    \renewcommand{\algorithmicrequire}{\textbf{Input:}}  
    \renewcommand{\algorithmicensure}{\textbf{Output:}} 
    \caption{The Overall Procedure of \model}  
    \label{alg:Framwork}
    \begin{algorithmic}[1]
        \REQUIRE An abnormal graph $\gG = (\bm{A}, \bm{X})$; Training epochs $T$; Balance parameters $\lambda_{1}$ and $\lambda_{2}$.
        \ENSURE An anomaly score list for the nodes.
        \FOR{$epoch \in 1,2,...T$}
        \STATE Obtain the attribute-specific, topology-specific, and consistency representation $\bm{H}^{(a)}=f_{a}(\bm{X})$, $\bm{H}^{(t)}=f_{t}(\bm{A})$, $\bm{H}^{(c)}=f_{gnn}(\bm{X},\bm{A})$, respectively.
        \STATE Calculate the correlation constraint loss $\gL_{cc}$ between any two of the three encoded representations ($\bm{H}^{(a)}$, $\bm{H}^{(t)}$, $\bm{H}^{(c)}$) via Eq.(\ref{eq:cc}).
        \STATE Obtain the cached attribute representation $\bm{Z}_{A}=\text{AC}(\bm{H}^{(a)}, \bm{H}^{(c)})$;
        \STATE Reconstruct the attribute matrix $\hat{\bm{X}}=g_{a}(\bm{Z}_{A}, \bm{A})$, and calculate the reconstruction loss $\gL_{attr}$ via Eq.(\ref{eq:attr});
        \STATE Obtain the cached structure representation $\bm{Z}_{T}=\text{AC}(\bm{H}^{(t)}, \bm{H}^{(c)})$;
        \STATE Reconstruct the adjacency matrix $\hat{\bm{A}}=g_{t}(\bm{Z}_{T})$, and calculate the reconstruction loss $\gL_{topo}$ via Eq.(\ref{eq:topo});
        \STATE Conduct the global consistency alignment with consistency representation $\bm{H}^{(c)}$, and calculate the reconstruction loss $\gL_{cons}$ via Eq.(\ref{eq:homo});
        \STATE Minimize the joint training loss $\gL = \lambda_{1}\cdot{\gL}_{attr} + (1-\lambda_{1})\cdot{\gL}_{topo} + \lambda_{2}\cdot{\gL}_{cons} + \gL_{cc}$;
        \STATE Update model's learnable parameters by using stochastic gradient descent;
        \ENDFOR
        \STATE Compute anomaly scores of nodes in the attributed network $\gG$ based on Eq.(\ref{eq:scoring}).
    \end{algorithmic}\label{alg:g3ad}
\end{algorithm}

\subsection{Complexity Analysis}
In this subsection, we conduct the complexity analysis of \model. First, the time complexity of encoding is $max(\gO(f_{a}(\cdot), f_{t}(\cdot), f_{gnn}(\cdot)))$. The complexity of $f_{a}(\cdot)$ and $f_{t}(\cdot)$ is $\gO(\lvert \gV \rvert d F)$, and the complexity of $f_{gnn}(\cdot)$ is $\gO((\lvert \gV \rvert + \lvert \gE \rvert) dF)$, where $F$ is the summation of all feature maps across different layers. Therefore, the complexity of the encoding guarding part is $\gO((\lvert \gV \rvert + \lvert \gE \rvert)dF)$.
Then, the complexity of the $\text{AC}(\cdot)$ module is $\gO(\lvert \gV \rvert dF)$, and the global consistency alignment can be processed simultaneously with $\gO(\lvert \gV \rvert dF)$.
Finally, the time complexity of the reconstruction modules is $max(\gO(g_{a}(\cdot), g_{t}(\cdot)))$, where $\gO(g_{a}(\cdot))$ is $\gO(\lvert \gE \rvert dF)$ and $\gO(g_{t}(\cdot))$ is $\gO(\lvert \gV \rvert^{2})$.
To sum up, the overall time complexity of \model is $\gO((\lvert \gV \rvert + \lvert \gE \rvert)dF+max(\lvert \gE \rvert dF, \lvert \gV \rvert^{2}))$.

%% file: experiment.tex
In this section, we conduct comprehensive experiments and in-depth analysis to demonstrate the effectiveness of \model. Specifically, we aim to answer the following research questions:
\textbf{RQ1:} How does \model perform compared with state-of-the-art models?
\textbf{RQ2:} How much do the architecture and components of \model contribute?
\textcolor{black}{\textbf{RQ3:} Can \model framework positively generalize in different GNN backbones?}
\textbf{RQ4:} How well does \model disentangled encode the information in each encoder? How does \model perform on different types of anomalies? \textcolor{black}{And how to explain the anomaly scoring effectiveness of \model?}
\textbf{RQ5:} How do key hyper-parameters impact \model's anomaly detection performance?

\subsection{Experimental Settings}
\subsubsection{Datasets}
We adopt six graph anomaly detection datasets on both synthetic and real-world scenarios that have been widely used in previous research~\cite{liu2021anomaly,chai2022can}, including four \textit{synthetic datasets}: \textbf{Cora}, \textbf{Citeseer}, \textbf{Pubmed}~\cite{sen2008collective}, and \textbf{Flickr}~\cite{tang2009relational}, and two \textit{real-world datasets}: \textbf{Weibo}~\cite{zhao2020error}, and \textbf{Reddit}~\cite{kumar2019predicting}.
The statistics are shown in Table \ref{tab:stats}.
The details of the datasets are introduced as follows.
\begin{itemize}[leftmargin=*]
    \item \textbf{Cora} \footnote{\url{https://linqs.soe.ucsc.edu/datac}\label{data_addr1}}~\cite{sen2008collective} is a classical citation network consisting of 2,708 scientific publications (contains 150 injected anomalies) along with 5,429 links between them. The text contents of each publication are treated as its attributes.
    \item \textbf{Citeseer} \textsuperscript{\ref{data_addr1}}~\cite{sen2008collective} is also a citation network consisting of 3,327 scientific publications (contains 150 injected anomalies) with 4,732 links. The node attribute in this dataset is defined the same as in the Cora dataset.
    \item \textbf{Pubmed}\textsuperscript{\ref{data_addr1}}~\cite{sen2008collective} is another citation network consisting of 19,717 scientific publications (contains 600 injected anomalies) with 44,338 links. The node attributes in this dataset are also defined as in the Cora dataset.
    \item \textbf{Flickr}\footnote{\url{http://socialcomputing.asu.edu/pages/datasets}}~\cite{tang2009relational} is a social network dataset acquired from the image hosting and sharing website Flickr. In this dataset, 7,575 nodes denote the users (contains 450 injected anomalies), 239,738 edges represent the following relationships between users, and node attributes of users are defined by their specified tags that reflect their interests on the website.
    \item \textbf{Weibo} \footnote{\url{https://github.com/zhao-tong/Graph-Anomaly-Loss}}~\cite{zhao2020error} is a user-posts-hashtag graph dataset from the Tencent-Weibo platform, which collects information from 8,405 platform users (contains 868 suspicious users) with 61,964 hashtags. The user-user graph provided by the author is used, which connects users who used the same hashtag.
    \item \textbf{Reddit} \footnote{\url{http://files.pushshift.io/reddit}}~\cite{kumar2019predicting} is a user-subreddit graph extracted from a social media platform, Reddit, which consists of one month of user posts on subreddits. The 1,000 most active subreddits and the 10,000 most active users (containing 366 banned users) are extracted as subreddit nodes and user nodes, respectively. We convert it to a user-user graph with 20,744,044 connections based on the co-interacted subreddit for our experiments.
\end{itemize}

\begin{table}[tbp]
  \centering
  \small
  \caption{Statistics of the experimental datasets.}
  \resizebox{\linewidth}{!}{
    \begin{tabular}{c|c|c|c|c}
    \toprule
    Dataset & \# nodes & \# edges & \# attributes & \# anomalies \\
    \midrule
    Cora  & 2,708 & 5,429 & 1,433 & 150 \\
    Citeseer & 3,327 & 4,732 & 3,703 & 150 \\
    Pubmed & 19,717 & 44,338 & 500 & 600 \\
    Flickr & 7,575 & 239,738 & 12,407 & 450 \\
    \midrule
    Weibo & 8,405 & 407,963 & 400 & 868 \\
    Reddit & 10,000 & 20,744,044 & 64 & 366 \\
    \bottomrule
    \end{tabular}%
  }
  \label{tab:stats}%
\end{table}%

For the five synthetic datasets, we adopted synthetic anomalies to validate models~\cite{ding2019deep,liu2021anomaly}.
Following the widely used anomaly injection approach in previous advances~\cite{liu2021anomaly,jin2021anemone,zheng2021generative,zhang2022subcr}, we inject a combined set of both topological and attributed anomalies for each experimental synthetic dataset with the following manner.
\begin{itemize}[leftmargin=*]
    \item \textbf{Injection of topological anomalies}. To obtain topological anomalies, the topological structure of networks is perturbed by generating small cliques composed of nodes that were originally not related. The insight is that in a small clique, a small group of nodes is significantly more interconnected with each other than the average, which can be considered a typical situation of topological anomalies in real-world graphs. Specifically, to create cliques, we begin by defining the clique size $p$ and the number of cliques $q$. When generating a clique, we randomly select $p$ nodes from the set of nodes $\gV$ and connect them fully. This implies that all the selected $p$ nodes are considered topological anomalies. To generate $q$ cliques, we repeat this process $q$ times. This results in a total of $p \times q$ topological anomalies. Following the previous works, the value of $p$ is fixed as 15, and the value of $q$ is set to 5, 5, 20, 20, 15 for Cora, Citeseer, Pubmed, ACM, and Flickr, respectively.
    \item \textbf{Injection of attributed anomalies}. We inject attributed anomalies by disturbing the attributes of nodes. To generate an attributed anomaly, a node $v_{i}$ is randomly selected as the target, and then another $k$ nodes $(v_{1}^{c}, ..., v_{k}^{c})$ are sampled as a candidate set $\gV^{c}$. Next, we compute the Euclidean distance between the attribute vector $\mathbf{x}_{c}$ of each $v^{c}\in\gV^{c}$ and the attribute vector $\mathbf{x}_{i}$ of $v_{i}$. We then select the node $v_{j}^{c}\in \gV^{c}$ that has the largest Euclidean distance to $v_{i}$ and change $\mathbf{x}_{i}$ to $\mathbf{x}_{j}^{c}$. Following the previous works, the value of $k$ is set to 50 in this paper.
\end{itemize}

\subsubsection{{Compared Baselines}}
We compare our proposed \model with twenty representative state-of-the-art models, which can be categorized into five main categories.

\noindent \textbf{Shallow detection methods:}
\begin{itemize}[leftmargin=*]
    \item \textbf{SCAN}~\cite{xu2007scan} is a classic clustering method that can be applied for anomaly detection, which clusters vertices based on structural similarity to detect anomalies.
    \item \textbf{MLPAE}~\cite{sakurada2014anomaly} utilizes autoencoders on both anomalous and benign data with shallow nonlinear dimensionality reduction on the node attribute.
\end{itemize}

\noindent \textbf{Enhanced graph neural networks:}
\begin{itemize}[leftmargin=*]
    \item \textbf{GAAN}~\cite{chen2020generative} is a generative adversarial framework with a graph encoder to obtain real graph nodes' representation and fake graph nodes' representation and a discriminator to recognize whether two connected nodes are from the real or fake graph.
    \item \textbf{ALARM}~\cite{peng2020deep} is a multi-view representation learning framework with multiple graph encoders and a well-designed aggregator between them.
    \item \textbf{AAGNN}~\cite{zhou2021subtractive} is an abnormality-aware graph neural network, which utilizes subtractive aggregation to represent each node as the deviation from its neighbors. 
\end{itemize}

\noindent \textbf{Graph autoencoder-based models:}
\begin{itemize}[leftmargin=*]
    \item \textbf{GCNAE}~\cite{kipf2016variational} is a classic variational graph autoencoder with the graph convolutional network as its backbone and utilizes the reconstruction loss for unsupervised anomaly detection.
    \item \textcolor{black}{\textbf{GATAE}~\cite{veličković2018graph} is a graph autoencoder based on graph attention networks with the same architecture as GCNAE.}
    \item \textbf{Dominant}~\cite{ding2019deep} is a deep graph autoencoder-based method with a shared encoder. It detects the anomalies by computing the weighted sum of reconstruction error terms of each node. 
    \item \textbf{AnomalyDAE}~\cite{fan2020anomalydae} is a dual graph autoencoder method based on the graph attention network, and the cross-modality interactions between network structure and node attribute are asymmetrically introduced on the node attribute reconstruction side.
    \item \textbf{ComGA}~\cite{luo2022comga} is a community-aware attributed graph anomaly detection framework with a designed tailored deep graph convolutional network.
    \item \textcolor{black}{\textbf{CoCo}~\cite{wang2024context} is a correlation-enhanced graph autoencoder with correlation analysis between the local and global contexts of each node.}
\end{itemize}

\noindent \textbf{Graph contrastive learning methods:}
\begin{itemize}[leftmargin=*]
    \item \textbf{CoLA}~\cite{liu2021anomaly} is a graph contrastive learning method. It detects anomalies by evaluating the agreement between each node and its neighboring subgraph sampled by the random walk-based algorithm.
    \item \textbf{ANEMONE}~\cite{jin2021anemone} is a multi-scale graph contrastive learning method, which captures the anomaly pattern by learning the agreements between node instances at the patch and context levels concurrently.
    \item \textbf{SL-GAD}~\cite{zheng2021generative} is a state-of-the-art anomaly detection model with generative and multi-view contrastive perspectives, which captures the anomalies from both the attribute and the structure space.
    \item \textbf{Sub-CR}~\cite{zhang2022subcr} is a self-supervised learning method that employs the graph diffusion-based multi-view contrastive learning along with attribute reconstruction.
    \item \textcolor{black}{\textbf{NLGAD}~\cite{duan2023normality} is a multi-scale graph contrastive learning network with a hybrid normality evaluation strategy.}
\end{itemize}

\noindent \textbf{Heterophily graph neural networks:}
\begin{itemize}[leftmargin=*]
    \item \textbf{MixHop}~\cite{abu2019mixhop} is a graph convolutional network with the mixed aggregation of multi-hop neighbors during a single message passing operation. We construct the unsupervised autoencoder architecture for it and the rest heterophily models to fit the unsupervised graph anomaly detections.
    \item \textbf{H2GCN}~\cite{zhu2020beyond} is a heterophily GNN by the separate encoding of ego \& neighbor embeddings with higher-order neighbors and intermediate representations.
    \item \textbf{LINKX}~\cite{lim2021large} is an MLP-based model for heterophily graph modeling, which separately embeds the adjacency and node features with simple MLP operations.
    \item \textbf{GloGNN}~\cite{li2022finding} is a method that considers both the homophily and heterophily properties on the graph with the combination of both low-pass and high-pass filters over the whole node set.
\end{itemize}

\begin{table*}[tbp]
  \centering
  \caption{Overall unsupervised anomaly detection comparison results on synthetic anomaly datasets (mean ± standard deviation in percentage over \textit{five} trial runs). The best and second-best results in each column are highlighted in \textbf{bold} font and \underline{underlined}.}
  \resizebox{\textwidth}{!}{
    \begin{tabular}{c|cc|cc|cc|cc}
    \toprule
    \multirow{3}[6]{*}{Model} & \multicolumn{8}{c}{Synthetic Datasets} \\
\cmidrule{2-9}          & \multicolumn{2}{c|}{Cora} & \multicolumn{2}{c|}{Citeseer} & \multicolumn{2}{c|}{Pubmed} & \multicolumn{2}{c}{Flickr} \\
\cmidrule{2-9}          & AUC   & \textcolor{black}{AP}    & AUC   & \textcolor{black}{AP}    & AUC   & \textcolor{black}{AP}    & AUC   & \textcolor{black}{AP} \\
    \midrule
    SCAN  & 0.6614±0.0140 & \textcolor{black}{0.0859±0.0044} & 0.6764±0.0063 & \textcolor{black}{0.0731±0.0032} & 0.7342±0.0037 & \textcolor{black}{0.1334±0.0032} & 0.6503±0.0120 & \textcolor{black}{0.3035±0.0227} \\
    MLPAE & 0.7560±0.0101 & \textcolor{black}{0.3528±0.0188} & 0.7404±0.0131 & \textcolor{black}{0.3124±0.0169} & 0.7477±0.0066 & \textcolor{black}{0.2508±0.0103} & 0.7466±0.0041 & \textcolor{black}{0.3484±0.0087} \\
    \midrule
    GAAN  & 0.7917±0.0118 & \textcolor{black}{0.3271±0.0124} & 0.8066±0.0036 & \textcolor{black}{0.3495±0.0101} & 0.7842±0.0042 & \textcolor{black}{0.0973±0.0018} & 0.7463±0.0043 & \textcolor{black}{0.3552±0.0119} \\
    ALARM & 0.8271±0.0223 & \textcolor{black}{0.2503±0.0379} & 0.8325±0.0121 & \textcolor{black}{0.3027±0.0559} & 0.8287±0.0044 & \textcolor{black}{0.1059±0.0020} & 0.6086±0.0034 & \textcolor{black}{0.0726±0.0013} \\
    AAGNN & 0.7590±0.0056 & \textcolor{black}{0.3744±0.0171} & 0.7202±0.0140 & \textcolor{black}{0.2345±0.0303} & 0.6693±0.0377 & \textcolor{black}{0.1210±0.0283} & 0.7454±0.0033 & \textcolor{black}{0.3506±0.0087} \\
    \midrule
    GCNAE & 0.7959±0.0104 & \textcolor{black}{0.3544±0.0350} & 0.7678±0.0114 & \textcolor{black}{0.3321±0.0243} & 0.7933±0.0081 & \textcolor{black}{0.2956±0.0083} & 0.7471±0.0056 & \textcolor{black}{0.3359±0.0125} \\
    \textcolor{black}{GATAE} & \textcolor{black}{0.8479±0.0098} & \textcolor{black}{0.3894±0.0239} & \textcolor{black}{0.8233±0.0126} & \textcolor{black}{0.2962±0.0219} & \textcolor{black}{0.8440±0.0061} & \textcolor{black}{0.3777±0.0103} & \textcolor{black}{0.7489±0.0050} & \textcolor{black}{0.3578±0.0111} \\
    Dominant & 0.8773±0.0134 & \textcolor{black}{0.3090±0.0438} & 0.8523±0.0051 & \textcolor{black}{0.3999±0.0092} & 0.8516±0.0034 & \textcolor{black}{0.1177±0.0019} & 0.6129±0.0035 & \textcolor{black}{0.0734±0.0013} \\
    AnomalyDAE & 0.8594±0.0068 & \textcolor{black}{0.3586±0.0148} & 0.8092±0.0059 & \textcolor{black}{0.3487±0.0103} & 0.7838±0.0042 & \textcolor{black}{0.0960±0.0016} & 0.7418±0.0051 & \textcolor{black}{0.3635±0.0107} \\
    ComGA & 0.7382±0.0162 & \textcolor{black}{0.1539±0.0242} & 0.7004±0.0150 & \textcolor{black}{0.0921±0.0085} & 0.7161±0.0065 & \textcolor{black}{0.0630±0.0015} & 0.6658±0.0033 & \textcolor{black}{0.1896±0.0120} \\
    \textcolor{black}{CoCo}  & \textcolor{black}{0.8642±0.0109} & \textcolor{black}{0.3938±0.0201} & \textcolor{black}{0.8336±0.0406} & \textcolor{black}{0.3175±0.0292} & \textcolor{black}{0.8053±0.0291} & \textcolor{black}{0.1826±0.0425} & \textcolor{black}{\underline{0.7590±0.0078}} & \textcolor{black}{\textbf{0.3683±0.0105}} \\
    \midrule
    CoLA  & 0.8866±0.0091 & \textcolor{black}{0.4638±0.0503} & 0.8112±0.0314 & \textcolor{black}{0.2355±0.0175} & \underline{0.9295±0.0047} & \textcolor{black}{\underline{0.4344±0.0134}} & 0.5612±0.0224 & \textcolor{black}{0.0694±0.0047} \\
    ANEMONE & \underline{0.8997±0.0048} & \textcolor{black}{\textbf{0.5417±0.0290}} & 0.8444±0.0178 & \textcolor{black}{0.3244±0.0361} & 0.9261±0.0080 & \textcolor{black}{0.4311±0.0313} & 0.5579±0.0271 & \textcolor{black}{0.0712±0.0062} \\
    SL-GAD & 0.8159±0.0238 & \textcolor{black}{0.3361±0.0337} & 0.7287±0.0201 & \textcolor{black}{0.2122±0.0273} & 0.8732±0.0066 & \textcolor{black}{0.3906±0.0313} & 0.7240±0.0087 & \textcolor{black}{0.3146±0.0123} \\
    Sub-CR & 0.8968±0.0118 & \textcolor{black}{0.4771±0.0102} & \underline{0.9060±0.0095} & \textcolor{black}{\underline{0.4751±0.0254}} & 0.9251±0.0041 & \textcolor{black}{\textbf{0.4936±0.0137}} & 0.7423±0.0038 & \textcolor{black}{0.3611±0.0145} \\
    \textcolor{black}{NLGAD} & \textcolor{black}{0.7369±0.0233} & \textcolor{black}{0.2308±0.0225} & \textcolor{black}{0.7559±0.0220} & \textcolor{black}{0.2177±0.0380} & \textcolor{black}{0.7331±0.0141} & \textcolor{black}{0.2862±0.0323} & \textcolor{black}{0.5045±0.0120} & \textcolor{black}{0.0676±0.0085} \\
    \midrule
    MixHop & 0.7796±0.0107 & \textcolor{black}{0.2412±0.0300} & 0.7401±0.0122 & \textcolor{black}{0.3146±0.0195} & 0.7726±0.0088 & \textcolor{black}{0.2212±0.0126} & 0.7447±0.0060 & \textcolor{black}{0.3517±0.0229} \\
    H2GCN & 0.7827±0.0104 & \textcolor{black}{0.3479±0.0285} & 0.7361±0.0169 & \textcolor{black}{0.3064±0.0139} & 0.7658±0.0065 & \textcolor{black}{0.2340±0.0037} & 0.7463±0.0042 & \textcolor{black}{0.3526±0.0135} \\
    LINKX & 0.7601±0.0098 & \textcolor{black}{0.3605±0.0242} & 0.7416±0.0117 & \textcolor{black}{0.3187±0.0180} & 0.7543±0.0070 & \textcolor{black}{0.2526±0.0103} & 0.7466±0.0042 & \textcolor{black}{\underline{0.3636±0.0065}} \\
    GloGNN & 0.7563±0.0083 & \textcolor{black}{0.3470±0.0269} & 0.7419±0.0124 & \textcolor{black}{0.3214±0.0220} & 0.7479±0.0067 & \textcolor{black}{0.2511±0.0103} & 0.7430±0.0013 & \textcolor{black}{0.3583±0.0054} \\
    \midrule
    G3AD (ours) & \textbf{0.9687±0.0013} & \textcolor{black}{\underline{0.4990±0.0191}} & \textbf{0.9688±0.0049} & \textcolor{black}{\textbf{0.5357±0.0115}} & \textbf{0.9321±0.0013} & \textcolor{black}{0.2179±0.0072} & \textbf{0.7693±0.0070} & \textcolor{black}{0.3270±0.0100} \\
    \bottomrule
    \end{tabular}%
    }
  \label{tab:main_syn}%
\end{table*}%

\subsubsection{Evaluation Metrics and Hyper-Parameter Settings}
We evaluate the models with Area under the ROC Curve (ROC-AUC) and Average Precision (AP), two widely-adopted metrics in previous works~\cite{bei2023reinforcement,jin2021anemone,ding2019deep}, to comprehensively evaluate the anomaly detection performance in different aspects. A higher AUC and AP value indicates better detection performance. Note that we run all experiments \textit{five} times with different random seeds and report the average results with standard deviation to prevent extreme cases.

\subsubsection{Hyper-parameter Settings} In the experiments on different datasets, the embedding size is fixed to 64, and the embedding parameters are initialized with the Xavier method~\cite{glorot2010understanding}. The loss function is optimized with Adam optimizer~\cite{kingma2014adam}. The learning rate of \model is searched from [$5\times 10^{\{-2,-3,-4\}}, 2\times 10^{\{-2,-3,-4\}}, 1\times 10^{\{-2,-3,-4\}}$]. \textcolor{black}{For the balance parameter of \model, we first tune $\lambda_1$ from [0, 1], then once $\lambda_1$ is determined, we select an appropriate $\lambda_2$ based on the fixed $\lambda_1$.}
For fair comparisons, all experiments are conducted on the CentOS system equipped with NVIDIA RTX-3090 GPUs.

\begin{table}[tbp]
  \centering
  \caption{Overall unsupervised anomaly detection comparison results on real-world anomaly datasets.}
  \resizebox{\linewidth}{!}{
    \begin{tabular}{c|cc|cc}
    \toprule
    \multirow{3}[6]{*}{Model} & \multicolumn{4}{c}{Real-world Datasets} \\
\cmidrule{2-5}          & \multicolumn{2}{c|}{Weibo} & \multicolumn{2}{c}{Reddit} \\
\cmidrule{2-5}          & AUC   & \textcolor{black}{AP}    & AUC   & \textcolor{black}{AP} \\
    \midrule
    SCAN  & 0.7011±0.0000 &  \textcolor{black}{0.1855±0.0000} & 0.4978±0.0000 & \textcolor{black}{0.0364±0.0000} \\
    MLPAE & 0.8946±0.0028 & \textcolor{black}{0.6696±0.0102} & 0.5108±0.0310 & \textcolor{black}{0.0359±0.0029} \\
    \midrule
    GAAN  & 0.9249±0.0000 & \textcolor{black}{\textbf{0.8104±0.0000}} & 0.5683±0.0001 & \textcolor{black}{0.0493±0.0001} \\
    ALARM & 0.9226±0.0000 & \textcolor{black}{0.8071±0.0000} & 0.5644±0.0003 & \textcolor{black}{0.0466±0.0001} \\
    AAGNN & 0.8066±0.0027 & \textcolor{black}{0.6679±0.0009} & 0.5442±0.0299 & \textcolor{black}{0.0400±0.0030} \\
    \midrule
    GCNAE & 0.8449±0.0032 & \textcolor{black}{0.5650±0.0030} & 0.5037±0.0015 & \textcolor{black}{0.0346±0.0001} \\
    \textcolor{black}{GATAE} & \textcolor{black}{\underline{0.9361±0.0079}} & \textcolor{black}{0.7848±0.0361} & \textcolor{black}{0.5643±0.0212} & \textcolor{black}{0.0448±0.0030} \\
    Dominant & 0.8423±0.0117 & \textcolor{black}{0.6163±0.0237} & \underline{0.5752±0.0056} & \textcolor{black}{\underline{0.0570±0.0019}} \\
    AnomalyDAE & 0.8881±0.0165 & \textcolor{black}{0.6681±0.0874} & 0.4315±0.0001 & \textcolor{black}{0.0319±0.0000} \\
    ComGA & 0.9248±0.0006 & \textcolor{black}{\underline{0.8097±0.0009}} & 0.4317±0.0001 & \textcolor{black}{0.0320±0.0001} \\
    \textcolor{black}{CoCo}  & \textcolor{black}{0.9257±0.0005} & \textcolor{black}{0.8095±0.0023} & \textcolor{black}{0.5388±0.0510} & \textcolor{black}{0.0403±0.0060} \\
    \midrule
    CoLA  & 0.4842±0.0238 & \textcolor{black}{0.1006±0.0155} & 0.5149±0.0233 & \textcolor{black}{0.0393±0.0020} \\
    ANEMONE & 0.3607±0.0120 & \textcolor{black}{0.0863±0.0147} & 0.4952±0.0188 & \textcolor{black}{0.0387±0.0026} \\
    SL-GAD & 0.4298±0.0073 & \textcolor{black}{0.0899±0.0036} & 0.5488±0.0142 & \textcolor{black}{0.0424±0.0032} \\
    Sub-CR & 0.6404±0.0070 & \textcolor{black}{0.4900±0.0103} & 0.5327±0.0156 & \textcolor{black}{0.0379±0.0013} \\
    \textcolor{black}{NLGAD} & \textcolor{black}{0.3259±0.0026} & \textcolor{black}{0.0792±0.0027} & \textcolor{black}{0.5631±0.0095} & \textcolor{black}{0.0433±0.0009} \\
    \midrule
    MixHop & 0.8612±0.0018 & \textcolor{black}{0.6278±0.0061} & 0.5400±0.0175 & \textcolor{black}{0.0398±0.0028} \\
    H2GCN & 0.8546±0.0020 & \textcolor{black}{0.5604±0.0088} & 0.5476±0.0113 & \textcolor{black}{0.0401±0.0010} \\
    LINKX & 0.8018±0.0094 & \textcolor{black}{0.2822±0.0088} & 0.5576±0.0162 & \textcolor{black}{0.0421±0.0020} \\
    GloGNN & 0.9128±0.0251 & \textcolor{black}{0.7465±0.1280} & 0.5436±0.0364 & \textcolor{black}{0.0468±0.0085} \\
    \midrule
    G3AD (ours) & \textbf{0.9514±0.0135} & \textcolor{black}{0.7434±0.0687} & \textbf{0.6207±0.0022} & \textcolor{black}{\textbf{0.0600±0.0008}} \\
    \bottomrule
    \end{tabular}%
    }
  \label{tab:main_realworld}%
\end{table}%

\subsection{Anomaly Detection Performance (RQ1)}
We first compare the main performance results between \model and the baseline models. \textcolor{black}{The performance comparison results on synthetic and real-world datasets are reported in Table \ref{tab:main_syn} and~\ref{tab:main_realworld}, respectively.}
From these results, we have the following observations:

\textbf{\model can achieve significant performance improvements over state-of-the-art methods.} 
\textcolor{black}{Specifically, for AUC metrics, \model achieves the best performance on all synthetic and real-world datasets among all baselines. For AP metrics, \model can also generally perform well and have the best or second-best performance on half of the datasets. Overall, \model has the top rank among all baselines on average in two metrics.}
The superior performance verifies the guarding schemes with correlation constraints, and the adaptive caching under the local and global anomaly detection tasks is able to help improve the anomaly distinguishability.

\textbf{Directly using the consistency GNNs or heterophily GNNs has sub-optimal detection performance.} We can find from the results that the performance of enhanced consistency-based GNNs (the second category) and heterophily-based GNNs (the fifth category) both have a certain gap with \model. Thus, it further verifies that, under such a disrupted phenomenon induced by unknown anomalies, \model can provide an effective way to guard the consistent homophily and utilize discrepant heterophily patterns.

\textbf{There are differences in the performance of baselines between synthetic and real-world datasets.} It can be found from the table that the graph contrastive learning-based methods are the best-performed baselines on synthetic datasets in general, while enhanced graph neural networks and graph autoencoders achieve better results on real-world datasets. One possible reason is that the contrastive objective function of these models is related one-fold, which is insufficient for directly applying to the irregular real-world datasets, and then enhanced graph neural networks and graph autoencoders.

\textbf{Anomaly detection on real-world datasets is significantly harder than on synthetic datasets.} The performance gap in real-world datasets between the models is large compared to synthetic datasets, especially the graph contrastive learning methods. This shows that the widely used anomaly injection scheme may lack diversity and be inadequate to simulate irregular patterns in the real world, \cite{liu2022bond} also has the same observation. Thus, we suggest that the model effectiveness examination should be conducted on both synthetic and real-world datasets for a more objective evaluation.

\subsection{Ablation Study (RQ2)}
In this subsection, we aim to conduct a fine-grained ablation study to analyze the contribution of components and architecture of our \model.
\subsubsection{Component Study}
To verify the effectiveness of the components, we conduct an ablation study on three variants: 
\textbf{\model-\textit{w/o AR}}, \textbf{\model-\textit{w/o TR}}, \textbf{\model-\textit{w/o CA}}, and their combinations, which remove the local attribute reconstruction, local topology reconstruction, and global consistency alignment, respectively.
From the results in Table \ref{tab:ablation1}, we can observe that \model notably outperforms all three variants and three ablation combinations in general. Specifically, first, the removal of attribute reconstruction and topology reconstruction has a greater impact, which shows that the modeling of graph properties is important for anomaly discrimination. Therefore, well-representing attributes and topology under anomaly discrepancy are significant for anomaly detection. Second, the effect of different modules varies from dataset, which is related to the diversity of anomaly definitions in different scenarios. Thus, the comprehensive consideration of topology-specific, attribute-specific, and consistent patterns is beneficial to the model.

\begin{table}[tbp]
  \centering
  \caption{\textcolor{black}{Component ablation study results on \model.}}
  \resizebox{\linewidth}{!}{
    \begin{tabular}{c|ccc}
    \toprule
    Variant & Flickr & Weibo & Reddit \\
    \midrule
    \model (ours) & \textbf{0.7693±0.0070} & \textbf{0.9514±0.0135} & \textbf{0.6207±0.0022} \\
    \midrule
    \model-\textit{w/o AR} & 0.5727±0.0047 & 0.7935±0.0887 & 0.6199±0.0007 \\
    \model-\textit{w/o TR} & 0.7466±0.0041 & 0.9258±0.0009 & 0.5136±0.0105 \\
    \model-\textit{w/o CA} & 0.7686±0.0074 & 0.9460±0.0108 & 0.6104±0.0186 \\
    \midrule
    \textcolor{black}{\model-\textit{w/o AR\&CA}} & \textcolor{black}{0.5722±0.0048} & \textcolor{black}{0.8307±0.1005} & \textcolor{black}{0.6199±0.0006} \\
    \textcolor{black}{\model-\textit{w/o TR\&CA}} & \textcolor{black}{0.7466±0.0040} & \textcolor{black}{0.9254±0.0008} & \textcolor{black}{0.5415±0.0321} \\
    \textcolor{black}{\model-\textit{w/o AR\&TR}} & \textcolor{black}{0.5533±0.0222} & \textcolor{black}{0.8058±0.0262} & \textcolor{black}{0.4218±0.0139} \\
    \bottomrule
    \end{tabular}%
    }
  \label{tab:ablation1}%
\end{table}%

\begin{figure}[tbp]
    \centering
    \includegraphics[width=\linewidth, trim=0cm 0cm 0cm 0cm,clip]{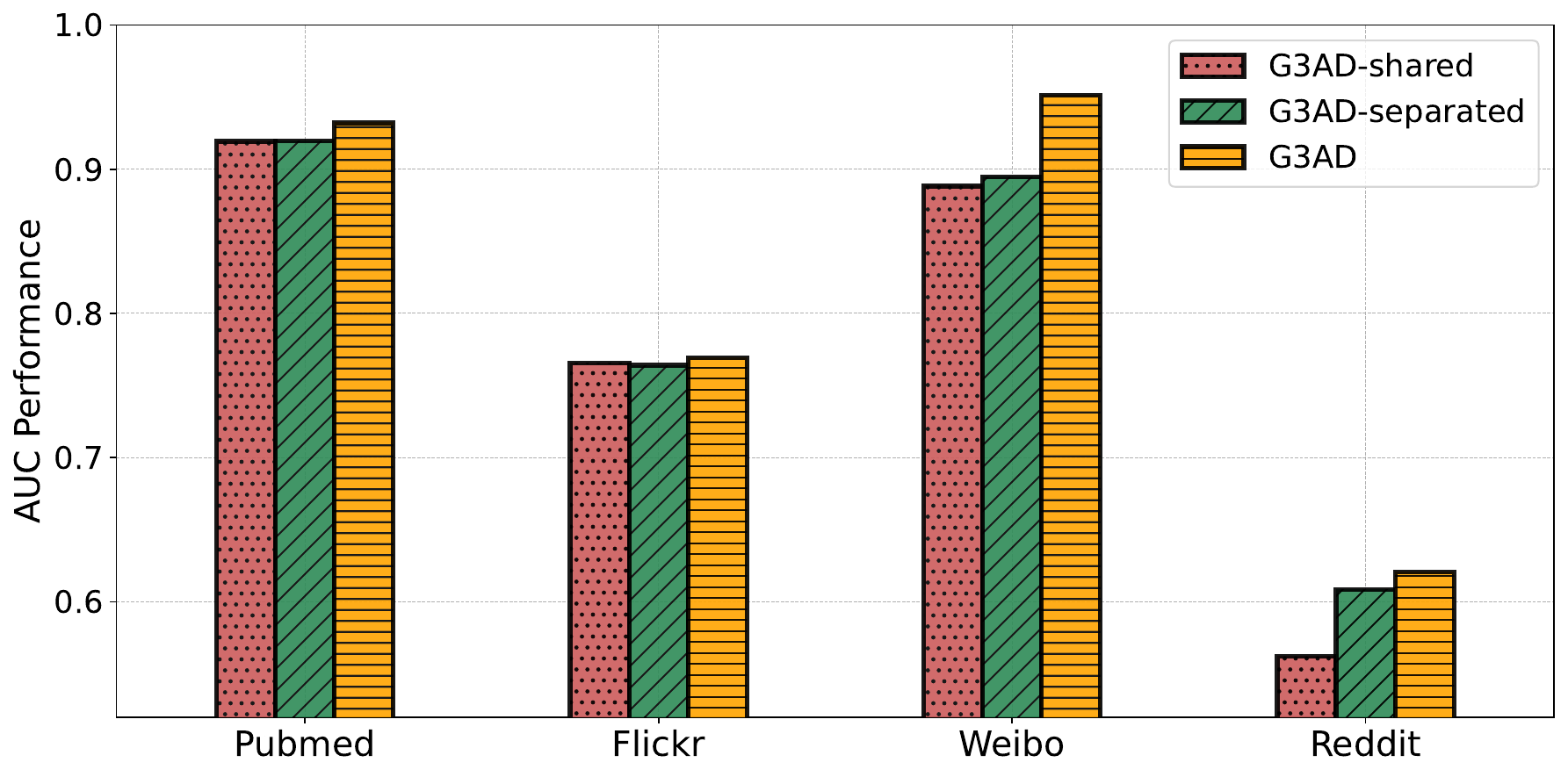}
    \caption{Architecture ablation study results on \model.}
    \label{fig:ablation_study}
\end{figure} 

\subsubsection{Architecture Study}
To study the effectiveness of \model's architecture, we further study the guarding architecture of \model with its two architecture variants:
\textbf{\model-\textit{shared}} removes the auxiliary encoders and shares all input information with a single GNN, which ignores the guarding of anomalies for the GNN.
\textbf{\model-\textit{separated}} replaces the three encoders with two parallel encoders without any information sharing, which is the simplest way without considering the consistent attribute-topology correlation that remained in the major normal nodes.
The results are shown in Figure \ref{fig:ablation_study}. From the results, we have the following observations: First, compared to \model-\textit{shared} variant, \model gains significant improvements, which proves the necessity of auxiliary guarding under abnormal graphs. Second, the improvements over \model-\textit{separated} demonstrate that simply separating attribute and topology information will miss the consistent patterns that remained in most normal nodes and lead to suboptimal performance. Therefore, the guarding architecture of \model is effective for unsupervised anomaly detection.

\subsection{Generalization Study (RQ3)}\label{sec:general_exp}
To investigate the generalization ability of the proposed \model framework in different GNN backbones, we evaluate the improvements of \model (w/ \model) for different GNN backbones (w/o \model). Specifically, we adopt the representative GCN~\cite{kipf2016semi}, GAT~\cite{veličković2018graph}, SAGE~\cite{hamilton2017inductive}, and GIN~\cite{xu2018gin} for \model in the GNN encoder, respectively.

\begin{figure}[tbp]
    \centering
    \includegraphics[width=\linewidth, trim=0cm 0cm 0cm 0cm,clip]{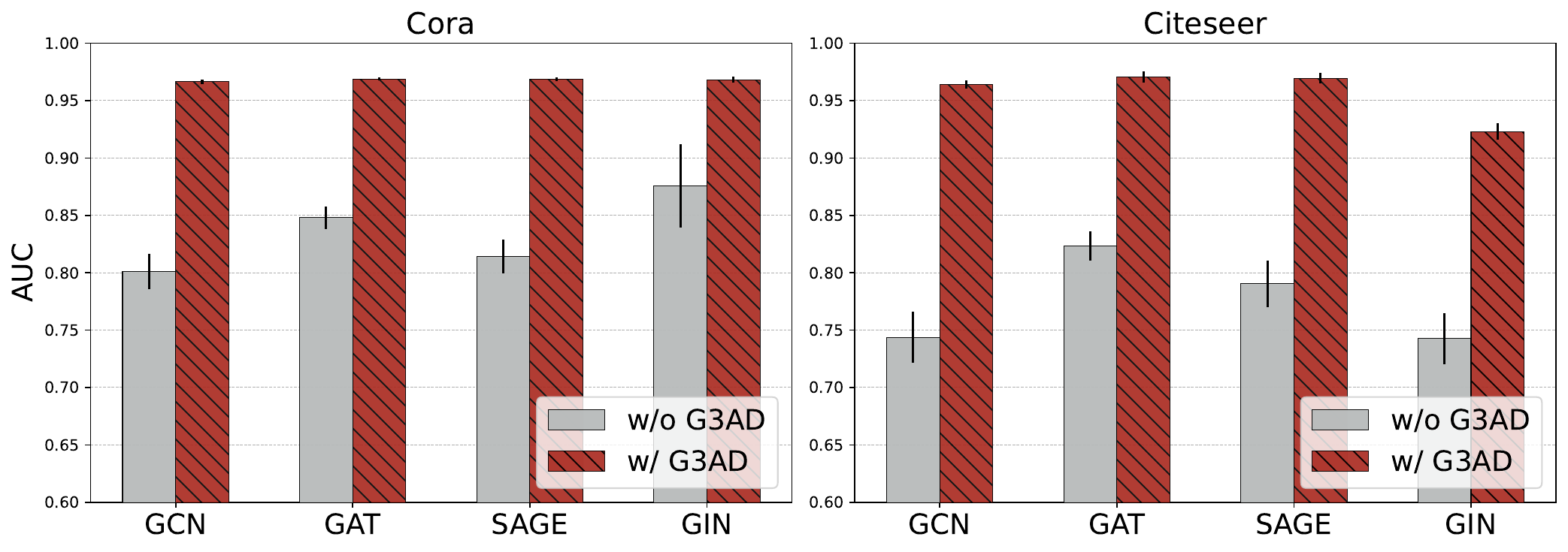}
    \caption{\textcolor{black}{Generalization study results of \model in different GNN backbones.}}
    \label{fig:generalize_study}
\end{figure}

The generalization study results are illustrated in Figure~\ref{fig:generalize_study}. From the results, we can find that \model can stably bring significant improvements across each GNN backbone. Specifically, equipped with our \model framework, the GNNs have 16.12\% and 23.62\% average AUC improvements on the Cora and Citeseer datasets, respectively. Furthermore, the improvements are evident across all GNN backbones. It has achieved at least a 10\% gain on each GNN, with an average improvement of 25.16\% specifically on GCN. These results highlight the generalization capability of G3AD in improving various GNN backbones across different datasets.

\begin{figure}[tbp]
    \centering
    \includegraphics[width=0.95\linewidth, trim=0cm 0cm 0cm 0cm,clip]{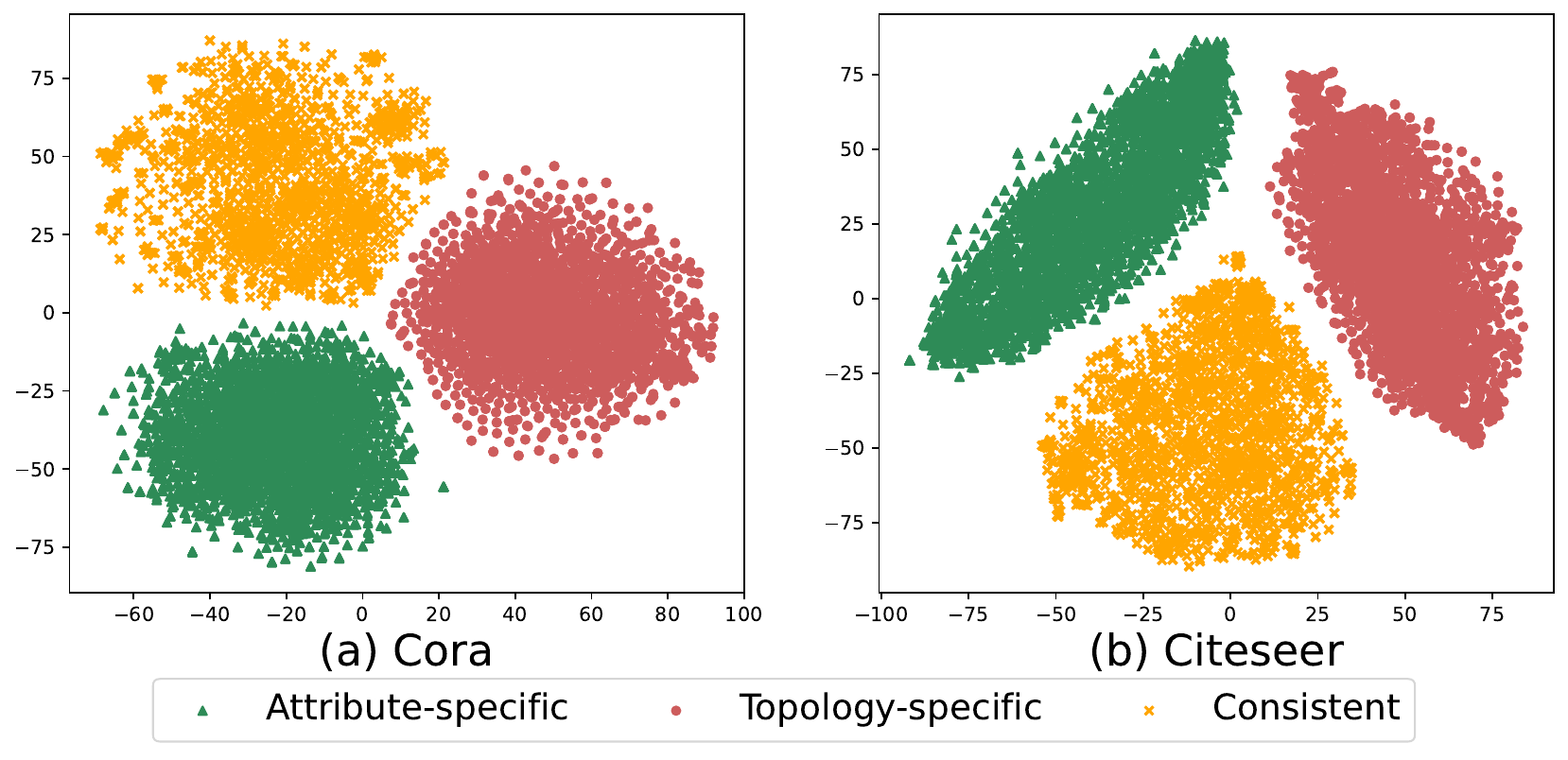}
    \caption{Visualization of the distribution of three aspects of embeddings after the \model encoding guarding.}
    \label{fig:emb_vis}
\end{figure}

\subsection{Case Study (RQ4)}
In this subsection, we aim to conduct case studies on \model with synthetic datasets to analyze the correlation-constrained GNN encoding guarding via embedding visualization and analyze the performance of \model on different types of anomalies.
\subsubsection{Embedding Visualization}
We first conduct the distribution visualization of attribute-specific, topology-specific, and consistent GNN embeddings after the correlation constrained guarding from the attribute encoder $f_{a}(\cdot)$, the topology encoder $f_{t}(\cdot)$, and the GNN encoder $f_{gnn}(\cdot)$ by reducing their dimension to two with the T-SNE~\cite{van2008visualizing} method.
The visualization results are illustrated in Figure \ref{fig:emb_vis}, which shows that the three aspects of representations are well distinguished from each other in the embedding space.
Therefore, it illustrates that the three aspects of representations have been well disentangled by \model to guard the GNN encoding.

\begin{table}[!t]
  \centering
  \caption{Case study of \model on attribute-induced, topology-induced, and mixed anomaly detection performance with top-3 performed baselines.}
  \resizebox{\linewidth}{!}{
    \begin{tabular}{cc|ccc}
    \toprule
    \multicolumn{2}{c|}{Anomaly} & Attribute & Topology & Mixed \\
    \midrule
    \multirow{4}[4]{*}{\rotatebox{90}{Cora}} &  CoLA  & 0.8458±0.0234 & 0.9273±0.0217  & 0.8866±0.0091  \\
    & ANEMONE & 0.8305±0.0151  & \underline{0.9688±0.0086}  & \underline{0.8997±0.0048}  \\
    & Sub-CR & \textbf{0.9806}±0.0042  & 0.8240±0.0328  & 0.8968±0.0118  \\
    \cmidrule{2-5} & \model (ours) & \underline{0.9623±0.0023}  & \textbf{0.9752±0.0013}  & \textbf{0.9687±0.0013} \\
    \midrule
    \multirow{4}[4]{*}{\rotatebox{90}{Citeseer}} &  CoLA  & 0.7437±0.0267  & 0.8785±0.0426  & 0.8112±0.0314   \\
    & ANEMONE & 0.7519±0.0220 & \underline{0.9368±0.0152}  & 0.8444±0.0178   \\
    & Sub-CR & \underline{0.9510±0.0099}  & 0.8733±0.0358  & \underline{0.9060±0.0095}  \\
    \cmidrule{2-5} & \model (ours) & \textbf{0.9637±0.0037}  & \textbf{0.9740±0.0073}  & \textbf{0.9688±0.0049}   \\
    \bottomrule
    \end{tabular}%
    }
  \label{tab:asm_result_new}%
\end{table}%

\begin{figure}[tbp]
    \centering
    \includegraphics[width=\linewidth, trim=0cm 0cm 0cm 0cm,clip]{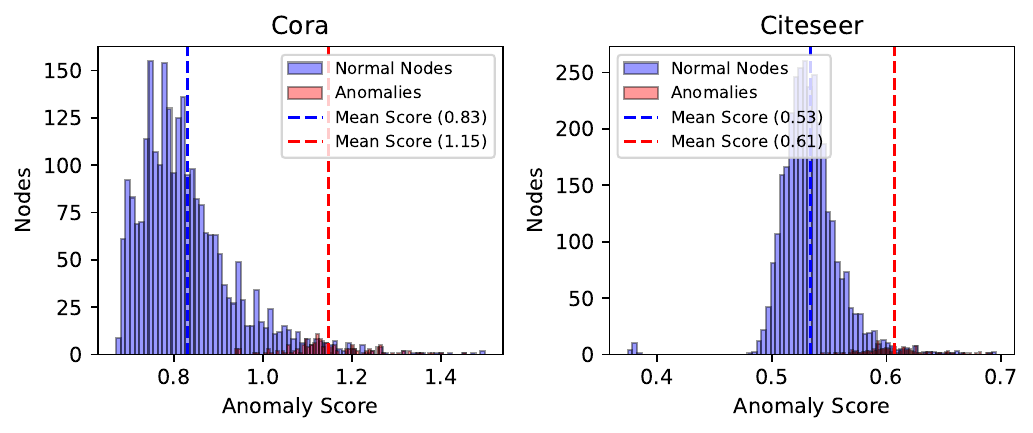}
    \caption{\textcolor{black}{Post-hoc analysis of anomaly score distribution predicted by \model.}}
    \label{fig:score_dist_vis}
\end{figure}

\subsubsection{Performance on Different Anomaly Types}
To further analyze different types of induced anomalies, we conduct a case study of \model compared with the overall top-3 performed baselines on two synthetic datasets. We report the detection performance on synthetic attributed anomalies, synthetic structure anomalies, and synthetic mixed (both attribute and structure perspectives) anomalies.
The split performance results are shown in Table \ref{tab:asm_result_new}. From the results, we find that the best baseline Sub-CR obviously tends to detect attribute anomalies, and the second-best performed model, ANEMONE, has an inclination towards detecting topology anomalies.
Our \model can generally balance and achieve good performance on all types of anomalies due to the explicit guarding of representation learning, guarding of directly abnormal graph reconstruction, and comprehensive consideration of different kinds of anomaly characteristics.

\subsubsection{\textcolor{black}{Anomaly Score Distribution}}
\textcolor{black}{To provide an explanation foundation for \model anomaly detection, we conduct a post hoc analysis by visualizing the anomaly score distribution predicted by \model. This distribution is crucial for both the gradient descent training phase and the anomaly discrimination process of the model. The visualization results are depicted in Figure~\ref{fig:score_dist_vis} for nodes in the Cora and Citeseer datasets. In the figure, we differentiate between normal nodes and anomalies using distinct colors: blue for normal nodes and red for anomalies. Additionally, we indicate the mean anomaly scores for both normal nodes and anomalies with dashed lines. From the results, the distinct separation between the normal and anomaly node distributions in both datasets shows the effectiveness of \model in anomaly detection. The clear distinction in anomaly scores allows for accurate identification and discrimination of anomalies.}

\begin{figure*}[t]
    \centering
    \includegraphics[width=0.97\linewidth, trim=0cm 0cm 0cm 0cm,clip]{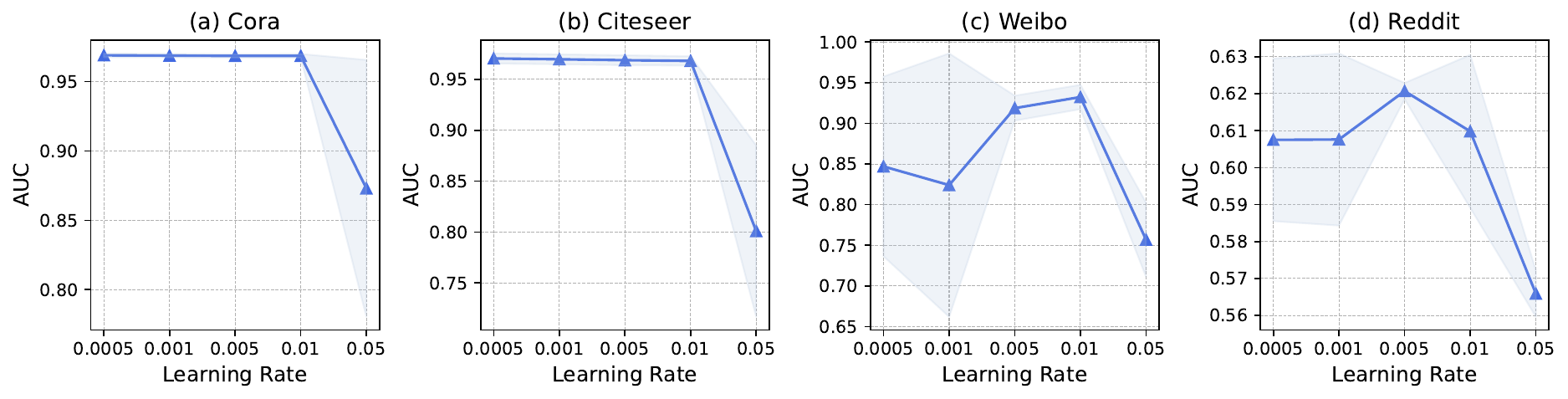}
    \vspace{-0.3em}
    \caption{Parameters study on \model with different learning rates.}
    \vspace{-0.3em}
    \label{fig:param_lr}
\end{figure*}

\begin{figure*}[t]
    \centering
    \includegraphics[width=0.97\linewidth, trim=0cm 0cm 0cm 0cm,clip]{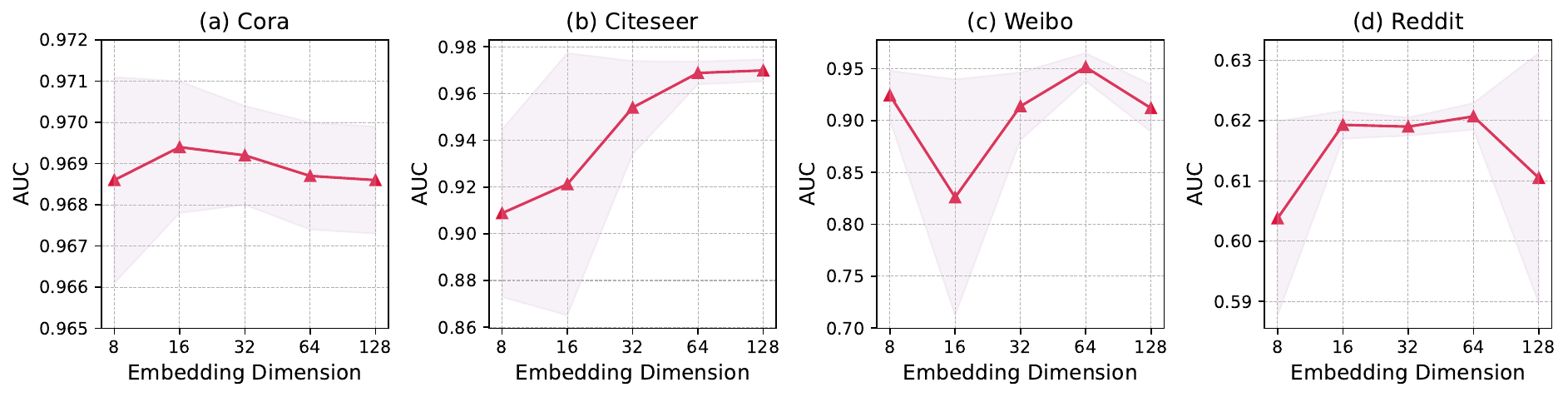}
     \vspace{-0.3em}
    \caption{Parameters study on \model with different embedding dimensions.}
    \vspace{-0.3em}
    \label{fig:param_emb}
\end{figure*}

\subsection{Parameter Analysis (RQ5)}
In this subsection, we aim to study the impact of different hyper-parameters on \model with two synthetic datasets and two real-world datasets, including the learning rate, the embedding dimension, the balanced parameters $\lambda_{1}$ and $\lambda_{2}$, and the readout function in Eq.(\ref{eq:readout_func}).

\subsubsection{Effectiveness of learning rate}
Then, the study of the effectiveness of the learning rate can be found in Figure \ref{fig:param_lr}. From the results, we can find that the performance of \model on synthetic datasets is less sensitive to changes in learning rate. As long as the learning rate is not set too large, the model's performance is similar under relatively small learning rates. For real-world datasets, the appropriate learning rate range will be smaller than that of synthetic datasets, but a learning rate value of around 0.005 can generally achieve relatively good anomaly detection performance.

\begin{figure}[t]
    \centering
    \includegraphics[width=0.975\linewidth, trim=0cm 0cm 0cm 0cm,clip]{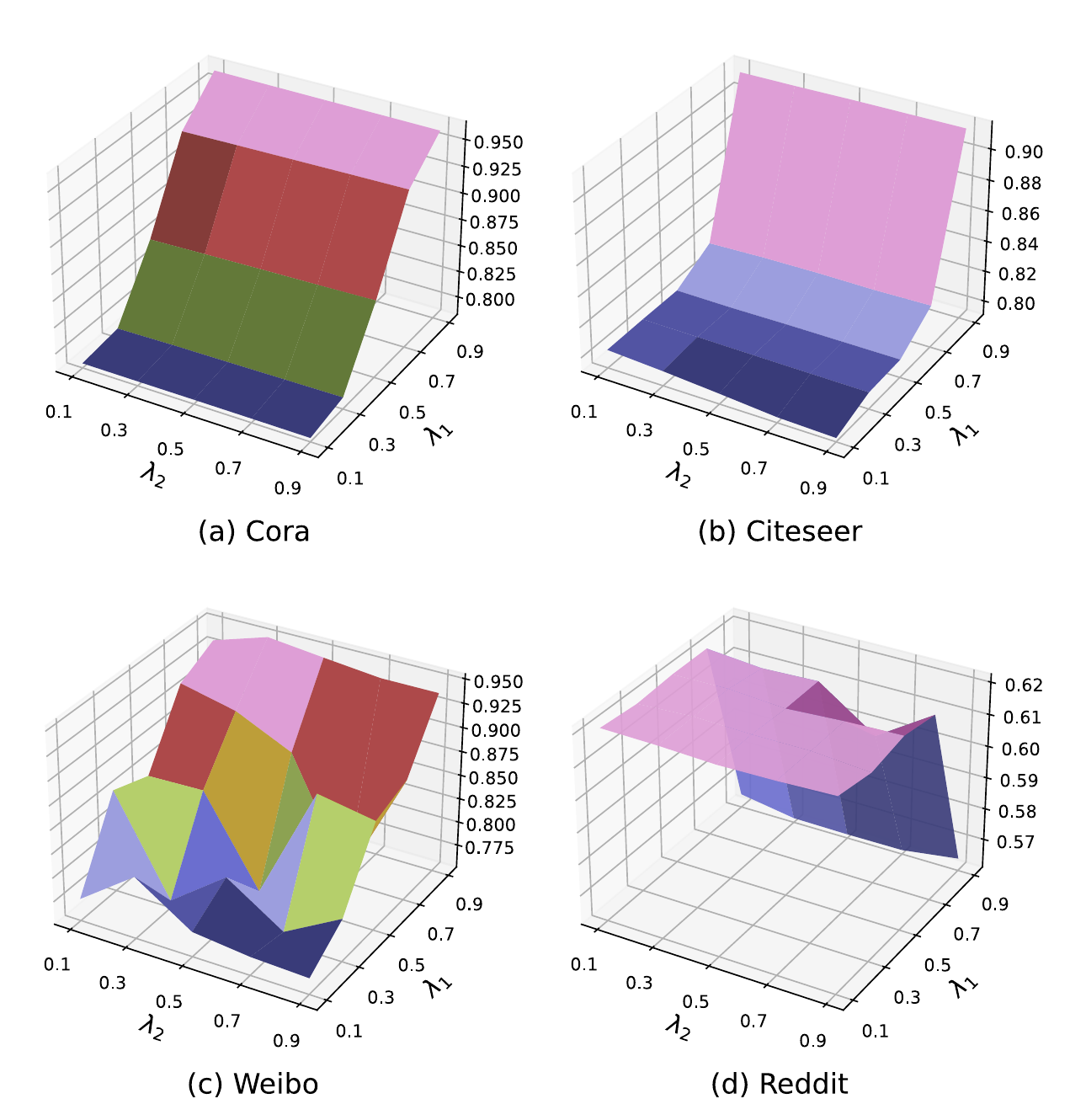}
    \vspace{-0.3em}
    \caption{Parameters study results with different combinations of $\lambda_{1}$ and $\lambda_{2}$.}
    \label{fig:heatmap_Cora_Citeseer}
\end{figure}

\subsubsection{Effectiveness of embedding dimension}
Furthermore, we have also analyzed the impact of different embedding dimensions on \model, as illustrated in Figure \ref{fig:param_emb}. It can be seen from the study results that, except for the Cora dataset, which can achieve optimal performance on a smaller 16-dimensional space, the other three datasets can achieve optimal performance on a 64-dimensional representation space in general, which is no need for us to continue to increase the representation dimension to achieve better performance.

\subsubsection{Effectiveness of balanced parameters}
We first investigate the effect of the different combinations of key balanced parameters $\lambda_{1}$ (balances the effect between topology and attribute reconstruction) and $\lambda_{2}$ (tunes the impact of consistency alignment) on \model. 
The parameter study results are shown in Figure~\ref{fig:heatmap_Cora_Citeseer} on the Cora, Citeseer, Weibo, and Reddit datasets. 
From the results, we can observe that the performance of \model varies with respect to balanced parameters $\lambda_{1}$ and $\lambda_{2}$. More specifically, first, changes in parameter $\lambda_{1}$ may bring more volatility than parameter $\lambda_{2}$. This is explainable since the consistency is related to both attribute and topology, which is more robust and stable~\cite{zhu2022does}. Furthermore, with a fixed $\lambda_{2}$ value, a relatively large value of $\lambda_{1}$ can make \model perform better, which means that the impact of the attribute reconstruction needs more attention and consideration.

\begin{table}[tbp]
  \centering
  \caption{Comparison results on different readout functions in the global consistency alignment.}
  \resizebox{\linewidth}{!}{
    \begin{tabular}{c|ccc}
    \toprule
    Readout & Flickr & Weibo & Reddit \\
    \midrule
    Mean  & \textbf{0.7693±0.0070} & \textbf{0.9514±0.0135} & \underline{0.6207±0.0022} \\
    Min   & 0.7679±0.0084 & \underline{0.9227±0.0422} & \textbf{0.6212±0.0017} \\
    Max   & 0.7682±0.0085 & 0.9075±0.0947 & 0.6194±0.0011 \\
    Attention & \underline{0.7684±0.0084} & 0.8664±0.1089 & 0.6186±0.0023 \\
    \bottomrule
    \end{tabular}%
    }
  \label{tab:readout}%
\end{table}%

\subsubsection{Effectiveness of readout function}
We further discuss the readout function selection of consistency alignment in Eq.(\ref{eq:readout_func}), which includes mean, min, max, and attention (scoring by an MLP) pooling as in Table \ref{tab:readout}. From the results, we can find that the simple non-parameter operation with mean pooling can achieve relatively good performance.
The possible reason is that due to the unbalanced number of normal and anomalies, simple mean pooling can reflect the global consistency from the representations of the majority of normal nodes.

%% file: conclusion.tex
In this paper, we study the problem of negative anomaly impact on GNNs in unsupervised graph anomaly detection, which is largely neglected by previous works. To address this issue, we propose \model, a simple but effective framework to guard GNNs against anomaly impacts in unsupervised settings. Specifically, \model introduces two guarding strategies with comprehensive anomaly detection perspectives. Firstly, \model guards the GNN encoder against encoding inconsistent information to enhance the node representation quality for anomaly distinguishing.
Then, we comprehensively include both local reconstruction and global alignment as the objectives for better detecting multiple anomalies. During this process, to guard the GNN-encoded representations against directly reconstructing the abnormal graph, \model further equips the representations with adaptive caching. 
Extensive experiments demonstrate that \model outperforms the state-of-the-art baselines.